\documentclass{ecai} 
\usepackage{latexsym}
\usepackage{amssymb}
\usepackage{amsmath}
\usepackage{amsthm}
\usepackage{booktabs}
\usepackage{enumitem}
\usepackage{graphicx}
\usepackage{color}
\usepackage{algorithmic}
\usepackage{algorithm}
\usepackage{array}
\usepackage[caption=false,font=footnotesize,labelfont=rm,textfont=rm]{subfig}
\usepackage{textcomp}
\usepackage{stfloats}
\usepackage{url}
\usepackage{verbatim}
\usepackage{ulem}
\usepackage{bm}
\counterwithout{table}{algorithm} 
\newcommand{\BibTeX}{B\kern-.05em{\sc i\kern-.025em b}\kern-.08em\TeX}
\begin{document}

%%%%%%%%%%%%%%%%%%%%%%%%%%%%%%%%%%%%%%%%%%%%%%%%%%%%%%%%%%%%%%%%%%%%%%%%

\begin{frontmatter}

%%% Use this command to specify your submission number.
%%% In doubleblind mode, it will be printed on the first page.

\paperid{0974} 

%%% Use this command to specify the title of your paper.

\title{Adversarial Topic-aware Prompt-tuning for Cross-topic Automated Essay Scoring}

%%% Use this combinations of commands to specify all authors of your 
%%% paper. Use \fnms{} and \snm{} to indicate everyone's first names 
%%% and surname. This will help the publisher with indexing the 
%%% proceedings. Please use a reasonable approximation in case your 
%%% name does not neatly split into "first names" and "surname".
%%% Specifying your ORCID digital identifier is optional. 
%%% Use the \thanks{} command to indicate one or more corresponding 
%%% authors and their email address(es). If so desired, you can specify
%%% author contributions using the \footnote{} command.

% \author[A]{\fnms{First}~\snm{Author}\orcid{....-....-....-....}\thanks{Corresponding Author. Email: somename@university.edu.}\footnote{Equal contribution.}}
% \author[B]{\fnms{Second}~\snm{Author}\orcid{....-....-....-....}\footnotemark}
% \author[B,C]{\fnms{Third}~\snm{Author}\orcid{....-....-....-....}} 

% \address[A]{Short Affiliation of First Author}
% \address[B]{Short Affiliation of Second Author and Third Author}
% \address[C]{Short Alternate Affiliation of Third Author}
\author[A]{\fnms{Chunyun}~\snm{Zhang}
% \orcid{0000-0002-3156-754X}
}
% \thanks{Corresponding Author. Email: somename@university.edu.}\footnote{Equal contribution.}}
\author[A]{\fnms{Hongyan}~\snm{Zhao}
% \orcid{0009-0006-4263-0811}\footnotemark
}
\author[A]{\fnms{Chaoran}~\snm{Cui}
% \orcid{0000-0003-3332-1348}
\thanks{Corresponding Author. Email: bruincui@gmail.com}
} 
\author[A]{\fnms{Qilong}~\snm{Song}
% \orcid{0009-0006-3143-7514}
}
\author[A]{\fnms{Zhiqing}~\snm{Lu}
% \orcid{0009-0004-8124-5479}
} 
\author[A]{\fnms{Shuai}~\snm{Gong}
% \orcid{0009-0008-2081-4841}
} 
\author[B]{\fnms{Kailin}~\snm{Liu}
% \orcid{0009-0008-2081-4841}
}
\address[A]{Shandong University of Finance and Economics}
\address[B]{University of Toronto}
% \address[C]{Short Alternate Affiliation of Third Author}

% hl834242117@gmail.com; bruincui@gmail.com; yangfujun@mail.sdufe.edu.cn; gsh8210@163.com; sdcjdxsql@163.com).}
% \thanks{K. Liu is with the Edward S. Rogers St. Department of Electrical \& Computer Engineering of University of Toronto, Toronto M5S 3G8, Canada (e-mail: klin.liu@mail.utoronto.ca).}

%%% Use this environment to include an abstract of your paper.

\begin{abstract}
Cross-topic automated essay scoring (AES) aims to develop a transferable model capable of effectively evaluating essays on a target topic. A significant challenge in this domain arises from the inherent discrepancies between topics. While existing methods predominantly focus on extracting topic-shared features through distribution alignment of source and target topics, they often neglect topic-specific features, limiting their ability to assess critical traits such as topic adherence. To address this limitation, we propose an Adversarial TOpic-aware Prompt-tuning (ATOP), a novel method that jointly learns topic-shared and topic-specific features to improve cross-topic AES. ATOP achieves this by optimizing a learnable topic-aware prompt—comprising both shared and specific components—to elicit relevant knowledge from pre-trained language models (PLMs). To enhance the robustness of topic-shared prompt learning and mitigate feature scale sensitivity introduced by topic alignment, we incorporate adversarial training within a unified regression and classification framework. In addition, we employ a neighbor-based classifier to model the local structure of essay representations and generate pseudo-labels for target-topic essays. These pseudo-labels are then used to guide the supervised learning of topic-specific prompts tailored to the target topic. Extensive experiments on the publicly available ASAP++ dataset demonstrate that ATOP significantly outperforms existing state-of-the-art methods in both holistic and multi-trait essay scoring. The implementation of our method is publicly available at: https://anonymous.4open.science/r/ATOP-A271.

\end{abstract}

\end{frontmatter}

%%%%%%%%%%%%%%%%%%%%%%%%%%%%%%%%%%%%%%%%%%%%%%%%%%%%%%%%%%%%%%%%%%%%%%%%
\section{Introduction}
% Automated Essay Scoring (AES) aims to automatically evaluate the quality of student essays written in response to various writing prompts, referred to as topics~\cite{ke2019automated}. In educational settings, the manual assessment of essays represents a significant time and resource burden for educators. Well-implemented AES systems address this challenge by automating the evaluation process, thereby allowing teachers to redirect their efforts toward other crucial teaching and learning activities. Beyond efficiency gains, AES technology delivers timely, consistent, and detailed feedback to students, fostering a more dynamic and effective learning cycle. Consequently, the ongoing development and enhancement of sophisticated AES systems is essential for advancing educational practices, particularly in large-scale academic settings and digital learning platforms where prompt, personalized feedback is highly valued.

Automated Essay Scoring (AES) aims to automatically evaluate the quality of student essays written in response to various prompts, referred to as topics~\cite{ke2019automated}. A well-designed AES system can alleviate the burden of manual grading for educators while providing timely and meaningful feedback to students.

% A well-designed AES system alleviates the burden of manual scoring for educators while providing students with timely and comprehensive feedback.
% AES is typically formulated as a regression task, with early approaches primarily relying on hand-crafted features~\cite{attali2006automated,klebanov2013using,vajjala2018automated}. Recently, deep learning has significantly advanced AES, leveraging its powerful representation learning capabilities to propose and achieve promising results~\cite{taghipour2016neural, dong2017attention, yang2020enhancing, sethi2022natural}.

AES is typically formulated as a regression task. Early approaches relied on hand-crafted features~\cite{attali2006automated,klebanov2013using,vajjala2018automated}, whereas recent advances in deep learning have enabled the use of powerful representation learning to significantly improve AES performance~\cite{taghipour2016neural, dong2017attention, yang2020enhancing, sethi2022natural}. However, these methods are typically topic-specific, which means that both training and test data are derived from the same topic. In real-world scenarios, however, obtaining a large number of annotated essays for a target topic is often impractical, while labeled data from other (source) topics may be abundant. This motivates the task of cross-topic AES, which seeks to develop a transferable model capable of effectively evaluating essays from a target topic~\cite{ke2019automated, ramesh2022automated}.

Intuitively, cross-topic AES requires capturing both topic-shared and topic-specific features to ensure comprehensive evaluation. 
% Topic-shared features, such as grammatical errors and coherence, enable a model to generalize and perform well across different topics. In contrast, topic-specific features, such as topic adherence, allow the model to achieve high accuracy and relevance for specific topics. 
Topic-shared features—such as grammar errors, coherence, and sentence structure—support generalization across topics. Topic-specific features—such as relevance to the topic or appropriate word choice—are critical for evaluating the content fidelity and topical alignment of an essay.  However, most existing cross-topic AES methods focus predominantly on topic-shared features. Some approaches use linguistic features (e.g., POS tags) in combination with hand-engineered topic-invariant features~\cite{ridley2020prompt,ridley2021automated, do2023prompt}, while others apply domain alignment techniques to match source and target distributions~\cite{li2020sednn,cao2020domain, chen2023pmaes,zhang2025pairwise}. These methods largely ignore topic-specific features, limiting their effectiveness in assessing traits like topic adherence.

% In recent years, pre-trained language models (PLMs) have acquired extensive knowledge through self-supervised training on large amounts of text~\cite{devlin2019bert,brown2020language,touvron2023llama}. To preserve the common knowledge embedded in these models, directly fine-tuning the parameters of the entire model for downstream tasks is impractical. Instead, the prompt-tuning method is employed, which involves freezing the parameters of the PLMs and inserting additional textual or learnable prompts to guide the PLMs~\cite{brown2020language, liu2021p}.  

With the emergence of pre-trained language models (PLMs) such as BERT~\cite{devlin2019bert}, GPT~\cite{brown2020language}, and LLaMA~\cite{touvron2023llama}, AES models can now leverage rich linguistic and contextual knowledge learned from large corpora. However, fine-tuning the entire PLM for downstream tasks risks overwriting this general knowledge and is computationally expensive. To address this, prompt-tuning has been introduced as a lightweight alternative, which freezes the PLM’s parameters and guides the model through additional learnable or textual prompts~\cite{brown2020language, liu2021p}.

% Motivated by this approach, we propose an \textbf{A}dversarial \textbf{TO}pic-aware \textbf{P}rompt-tuning (\textbf{ATOP}) method, which approaches the joint learning of topic-shared and topic-specific features as an  optimization of topic-shared and topic-specific soft prompts. By optimizing these prompts, we can guide the cross-topic AES task's topic-shared and topic-specific features encoded in PLMs. Specifically, topic-shared and topic-specific prompts are prepended to each essay, forming an enhanced input that is processed by a PLM. The output at the [CLS] position is then used as the feature representation of the essay. ATOP jointly learns topic-shared and topic-specific prompts within a unified framework that combines regression and classification tasks for the AES task.

Motivated by this paradigm, we propose an \textbf{A}dversarial \textbf{TO}pic-aware \textbf{P}rompt-tuning (\textbf{ATOP}) method, which formulates the learning of topic-shared and topic-specific features as a joint optimization of corresponding soft prompts within a unified modeling framework. By optimizing these prompts, ATOP effectively guides pre-trained language models (PLMs) to encode both topic-shared and topic-specific information for cross-topic AES. Specifically, topic-shared and topic-specific prompts are prepended to each essay, forming an enriched input that is fed into a PLM. The resulting representation at the [CLS] token is used as the essay’s feature representation. ATOP integrates the learning of both prompt types through a joint framework that combines regression and classification objectives for the AES task.

To learn the topic-shared prompt, we introduce adversarial training via a min-max game between a discriminator and the topic-shared prompt. The discriminator aims to distinguish whether a prompt-enhanced essay originates from the source or target topic, while the prompt is optimized to confuse the discriminator. However, as noted in~\cite{chen2021representation}, aligning deep representations can alter feature scales, posing challenges for regression tasks, which are sensitive to such changes. To address this issue, we jointly model AES as a regression and classification task, applying adversarial training only within the classification branch to enhance robustness to feature scaling.

To learn the topic-specific prompt, we adopt a pseudo-labeling strategy using a neighbor-based classifier. This classifier captures the local structure of representations stored in a memory bank to generate pseudo-labels for target-topic essays. These pseudo-labeled samples, combined with labeled essays from source topics, are used to supervise the learning of topic-specific prompts for each topic.

In summary, the contributions of this paper are as follows:
\begin{itemize}
\item We propose a novel cross-topic AES method in a prompt-tuning paradigm, which optimizes topic-shared and topic-specific prompts to guide the corresponding features in PLMs. To our knowledge, this is the first work to introduce prompt-tuning into cross-topic AES.

% \item We propose a novel approach to learn the robust topic-shared prompt by applying adversarial training within a joint classification and regression framework, specifically leveraging the classification component to mitigate the feature scale sensitivity inherent in the regression task.

% \item We design a robust adversarial training strategy to learn topic-shared prompts within a joint classification and regression framework, leveraging the classification branch to mitigate feature scale sensitivity in the regression task.
\item We design a robust adversarial training strategy to learn topic-shared prompts within a joint classification and regression framework by integrating adversarial training into the classification branch, thereby mitigating the feature scale sensitivity inherent in the regression task.

% \item We introduce adversarial training in a joint modeling framework of regression and classification to learn the topic-shared prompt, enhancing the model's robustness to feature scale variations.
% \item We jointly model AES as both a regression and classification task, incorporating adversarial training within the classification framework to align source and target topics. This approach mitigates the adverse effects of feature scaling caused by domain alignment, improving the model's robustness.
% \item We jointly model AES as both regression and classification tasks, which integrates domain adversarial training with classification tasks to learn topic-shared prompts. This approach can avoid the adverse effects of changing feature scales on regression tasks during the adversarial process.
\item Extensive experiments on the public ASAP++ dataset demonstrate that ATOP outperforms nine state-of-the-art baselines on both holistic and multi-trait scoring tasks, particularly excelling in traits related to topic-specific characteristics.
\end{itemize}

\section{Related Work}
\subsection{Automated Essay Scoring}
Early AES methods \cite{attali2006automated,klebanov2013using,vajjala2018automated} focused on handcrafted features. Recently, deep learning has significantly advanced AES and achieve promising results. Deep AES methods can be categorized into topic-specific and cross-topic methods.
% Early approaches to AES~\cite{larkey1998automatic,shermis2003automated,uto2020neural} focused on combining traditional machine learning algorithms with expert-designed manual features, which have good interpretability but require significant human resources. Recently, deep neural network-based methods have been widely applied to AES tasks with promising results, and these methods can be categorized into topic-specific and cross-topic methods.

\subsubsection{Topic-specific Method}
Topic-specific AES aims to develop an automated scoring model trained and tested on essays from the same topic. From early methods using classic deep neural networks~\cite{taghipour2016neural, tay2018skipflow} to those based on attention mechanisms~\cite{dong2017attention} and pre-trained models~\cite{rodriguez2019language,yang2020enhancing, mayfield2020should, sethi2022natural}, the performance of topic-specific AES has significantly improved.

\subsubsection{Cross-topic method}
% Cross-topic AES aims to develop a transferable automatic scoring model that can effectively evaluate essays on a target topic. Some of these methods utilize common language features  as input, such as parts of speech (POS), and combined them with hand-designed topic-agnostic features~\cite{ridley2020prompt, ridley2021automated, do2023prompt}, others learn topic-agnostic features by aligning the target topic with source topics through two-stage methods~\cite{jin2018tdnn,li2020sednn,cao2020domain, song2020multi} and topic mapping methods~\cite{chen2023pmaes}. Additionally, Jiang et al.~\cite{jiang2023improving}  and Chen et al ~\cite{chen2024plaes} proposed models based on disentangled representation learning and meta learning, respectively, to address topic generalization. 

Cross-topic method aims to develop a transferable automated scoring model that can effectively evaluate essays on a target topic. Some of these methods utilize common language features  as input, such as POS, and combine them with hand-designed topic-agnostic features. Ridley et al.~\cite{ridley2020prompt} used POS feature as input and combined them with hand-designed topic-agnostic features to enhance model performance. They subsequently extended this approach to achieve cross-topic trait scoring (CTS) ~\cite{ridley2021automated}, providing students with multi-trait feedback. Do et al.~\cite{do2023prompt} incorporated prompt information into the CTS model to conduct comprehensive cross-topic AES.  Others learn topic-agnostic features by aligning the target topic with source topics through two-stage methods. Jin et al.~\cite{jin2018tdnn} proposed a model that first trained a topic-agnostic model to generate pseudo-labels for target topic essays. Subsequently, a scoring model for the target topic was trained using essays with the highest and lowest pseudo-label scores. Building on this, Li et al.~\cite{li2020sednn} improved cross-topic AES performance by using an adversarial network to learn a topic-independent model initially and adding topic information in the subsequent stage. Cao et al.~\cite{cao2020domain} constructed two topic-agnostic self-supervised tasks to learn shared features between topics, then used adversarial training to score target topic essays. Chen et al. \cite{chen2023pmaes} employed a contrastive learning-based technique to align features between source and target topics within the CTS model. Additionally, Jiang et al.~\cite{jiang2023improving}  and Chen et al. ~\cite{chen2024plaes} proposed models based on disentangled representation learning and meta-learning, respectively, to address topic generalization.

\subsection{Prompt-tuning}

Prompt-tuning methods adapt PLMs to downstream tasks by freezing the parameters of the PLMs and adjusting a set of trainable prompts~\cite{brown2020language}. Early prompt-tuning methods used manually defined textual templates (hard prompts) for tasks such as classification and natural language inference~\cite{schick2021exploiting, wu2022adversarial}, but these prompts led to significant variations in model performance. Consequently, recent methods have focused on using learnable prompts, such as soft prompts~\cite{lester2021power}, prefix-tuning~\cite{li2021prefix}, and P-tuning V2~\cite{liu2021p}, to better adapt PLMs to downstream tasks. 

\section{Method}
\begin{figure*}[ht]
\centering
 \centering
    \includegraphics[width=0.9\linewidth]{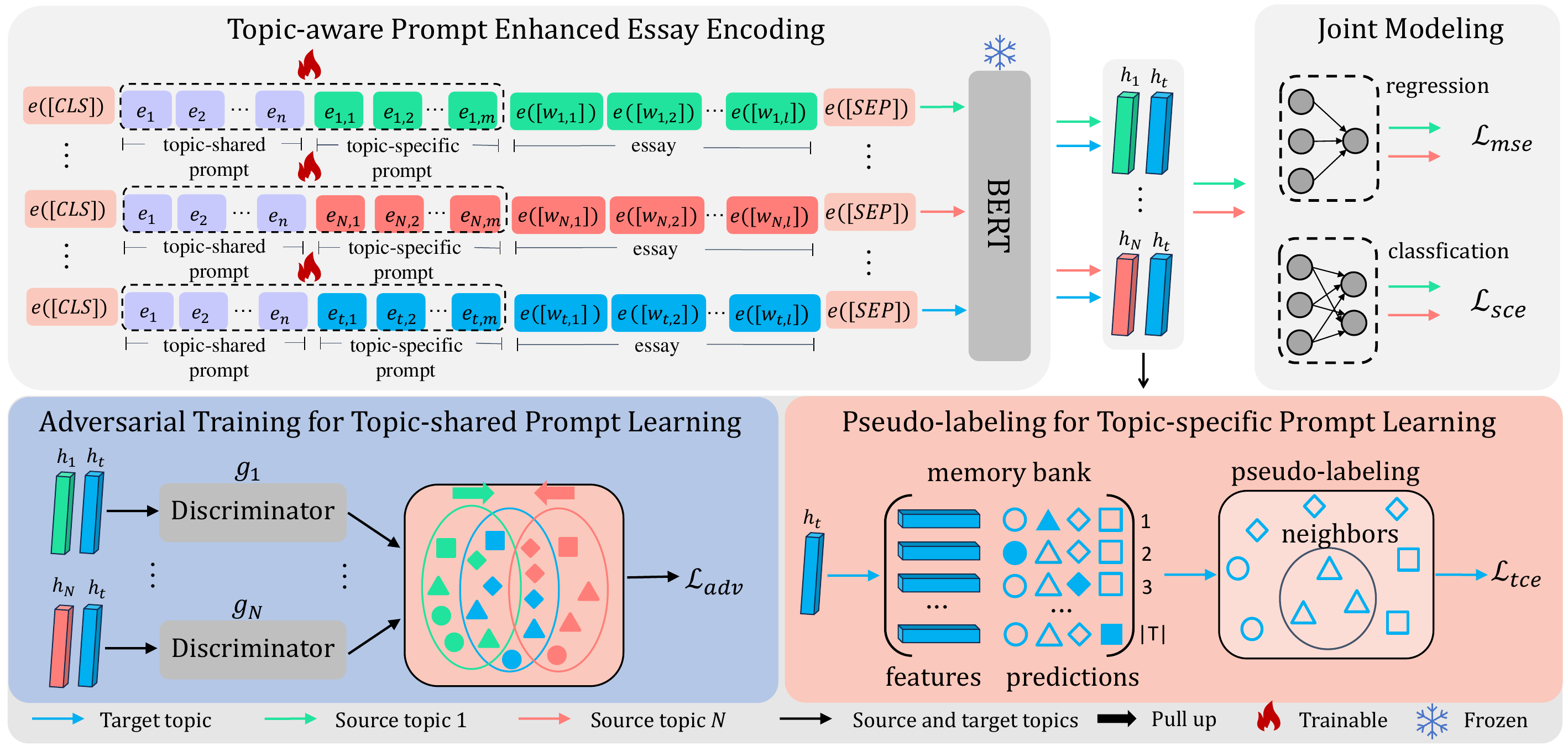}
\caption{The framework of ATOP. ATOP optimizes topic-shared and topic-specific prompts within a joint classification and regression framework. }\label{fig:framework}
\vspace{-2mm}
\end{figure*}

In this paper, we propose an Adversarial TOpic-aware Prompt-tuning (ATOP) method for cross-topic AES. ATOP optimizes topic-shared and topic-specific prompts within a joint classification and regression framework. Adversarial training  between the topic-shared prompt and the topic discriminator is employed to learn the topic-specific prompt. Pseudo-labeling method is also used to assign pseudo-labels for target topic, aiding in the learning of the topic-specific prompt.  The overall framework of ATOP is illustrated in Figure \ref{fig:framework}.

% In this paper, we propose an Adversarial TOpic-aware Prompt-tuning (ATOP) method for cross-topic AES. ATOP incorporates topic-specific knowledge into learnable topic-aware prompts, encompassing both topic-agnostic and topic-specific prompts, to guide the PLM within a joint classification and regression framework. Adversarial training is employed between the topic-agnostic prompt and the topic discriminator to learn the topic-specific prompt. The overall framework of ATOP is illustrated in Figure \ref{fig:framework}.

% detail the soft prompt tuning framework, which uses topic-agnostic and topic-specific soft prompt to motivate the PLM capture both topic-agnostic features and topic-related features. In addition, we introduce pairwise adversarial training method and pseudo-labeling method to better learn topic-agnostic topic-related features of the target domain, respectively. Finally, we detail the training process and the joint learning method based on classification and regression tasks. Our model framework is shown in Fig. 2.

\subsection{Problem Definition}
In this paper, we jointly model AES as both regression and classification tasks to predict the holistic and multi-trait scores for essays in a cross-topic setting. 

Formally, suppose there are scored essays from $N$ source topics $\bm{S_{}} = \{ ({\bm{X}_i},{\bm{Y}_i^{}, \bm{C}_i^{})\}} _{i = 1}^N$ and unscored assays from a target topic $\bm{T} = \{ \bm{x}_t^{}\} _{t = 1}^{|\bm {T}|}$. ${\bm{X}_i} = \{ x_s^i\}_{s=1}^{|{\bm{S}_i}|}$ denotes the scored essays from the $i$th source topic, $\bm{Y}_i^{} = \{ \bm{y}_s^i\} _{s=1}^{|{\bm{S}_i}|}$ denotes the corresponding scores of ${\bm{X}_i}$. Here, we uniformly label the holistic score as one of traits. Hence, $\bm{y}_s^i = \{ y_{s,1}^{i},y_{s,2}^{i},...,y_{s,K}^{i}\}$, where $y_{s,1}$ corresponds to the holistic score and the other ones correspond to the $K-1$ trait scores. Note that, to model AES as a classification task, we map scores $\bm{Y}_i$ into $\rm{C}$ levels (i.e. excellent, good, moderate, and poor) according to different score intervals (see details in Appendix A), and $\bm{C}_i^{} = \{ c_s^i\} _{s=1}^{|{\bm{S}_i}|}$, $c_s^{} \in \{ 0,1,2,...,\rm{C}- 1\}$ denotes the corresponding  categories of ${\bm{X}_i}$. Our objective is to train a transferable model $\phi$ based on $\bm{S}$ and $\bm{T}$. Then, we can predict credible holistic and trait scores for essay $x_t$ from $\bm{T}$ by: 
\begin{equation}
  \bm{\hat y}  = \phi(\bm{S},\bm{T},x_t;\theta ),  
\end{equation}
where $\theta $ is the trainable parameter of the model.

\subsection{Topic-aware Prompt Enhanced Essay Encoding}

In this paper, we introduce prompt-tuning into cross-topic AES by conceptualizing it as a task of eliciting relevant knowledge from PLMs. Unlike existing prompt-tuning methods that only utilize domain-shared prompts~\cite{wu2022adversarial}, we incorporate topic-specific information into a topic-aware prompt, which consists of both topic-shared and topic-specific components. 
% The topic-agnostic prompt guides the extraction of topic-shared knowledge from PLMs, while the topic-specific prompt targets the extraction of topic-dependent knowledge.
% Most existing prompt-tuning methods utilize domain-agnostic prompts, potentially limiting their effectiveness in addressing domain shift issues in cross-topic AES. In this work, we incorporate topic-specific information into learnable, topic-aware prompt, which consists of both topic-agnostic and topic-specific components. The topic-agnostic prompt is designed to guide the extraction of topic-shared knowledge from PLMs, while the topic-specific prompt targets the extraction of topic-dependent knowledge.

% Motivated by [16], we first insert before each essay with the learnable topic-shared and topic-specific soft prompts. Then, we input the essay with prompt into a PLM and view the output of the \textit{[CLS]} token of the PLM as the feature representation of the essay.

Specifically, we define a topic-shared prompt as  ${\bm{p}_{{\text{shared}}}} = \{ {\bm{e}_1},{\bm{e}_2},...,{\bm{e}_n}\} $ with $n$ learnable topic-shared vectors for all source and target topics. For each topic $i$, we further define a topic-specific prompt as ${\bm{p}_{{\text{specific}}}^i} = \{ {\bm{e}_{i,1}},{\bm{e}_{i,2}},...,{\bm{e}_{i,m}}\} $ with $m$ learnable topic-specific vectors. Given an essay ${x} = \{ {w_1},{w_2},...,{w_l}\} $, where $l$ denotes its length, we obtain its prompt enhanced embedding as:
% Specifically, given an essay $\bm{x} = \{ {w_1},{w_2},...,{w_l}\} $, where $l$ denotes its length. A template function $T(\cdot )$  converts the input sequence $\bm{x}$ into token sequence ${\bm{x}_{prompt}} = T(\bm{x})$, which contains embedding of the input sequence, $n$ learnable topic-agnostic vectors ${\bm{p}_{{\rm{agnostic}}}} = \{ {\bm{e}_1},{\bm{e}_2},...,{\bm{e}_n}\} $ and $m$ learnable topic-specific vectors ${\bm{p}_{{\rm{specific}}}} = \{ {\bm{e}_1},{\bm{e}_2},...,{\bm{e}_m}\} $, two embedding of special tokens of \<SEP\> and \<CLS\>. The input of the pre-trained model $M$ is denoted as:
% \begin{equation}
% \begin{split}
% {\bm{x}} = &[e([CLS]),{\bm{p}_{shared}},{\bm{p}_{specific}^i},e(\bm{x}),e([SEP])] ,
% \end{split}
% \end{equation}
\begin{equation}
\bm{x} = [e(\text{[CLS]}),{\bm{p}_{\text{shared}}},{\bm{p}_{\text{specific}}^i},e(\bm{x}),e(\text{[SEP]})],
\end{equation}
where {[CLS]} and {[SEP]} represents the beginning and the ending symbol, $e(\cdot)$ denotes the embedding operation of the PLM $\mathcal{M}$, which maps the input essay into a pre-trained word embedding sequence. We then input $\bm{x}$ into the PLM and obtain the output of the [CLS] position as $ {\bm{h}_{\text{[CLS]}}} = \mathcal{M}({\bm{x}})$. Here, ${\bm{h}_{\text{[CLS]}}} \in {\mathbb{R}^d}$, $d$ denotes the dimension of the output of $\mathcal M$. Following~\cite{liu2021p},  we use the $\bm{h}_{\text{[CLS]}}$ as the representation of the essay $x$.
% , subsequent modules are implemented based on ${\bm{h}_{[mask]}}$.
\subsection{Joint Modeling}
Unlike existing AES approaches that frame the task exclusively as either regression or classification, we formulate it as a joint regression-classification problem. By learning shared representations across both tasks, we enable mutual performance enhancement. Furthermore, we incorporate adversarial training grounded in the classification objective to optimize a topic-shared prompt, thereby enhancing the model's robustness to feature scaling induced by domain alignment.
% Unlike existing AES methods that model AES solely as a classification or regression task, we jointly model AES as the regression and classification tasks. This joint modeling approach allows for the learning of shared features between the two tasks, leading to a mutual improvement in their performance. In addition, domain adversarial training has proven highly effective in domain adaptation by learning domain-invariant feature representations through a min-max game between a domain discriminator and a feature extractor. However, as highlighted by [16], aligning the feature representations of two domains inevitably alters the feature scale. While classification tasks demonstrate robustness to variations in feature scale, regression tasks do not exhibit the same resilience. Consequently, the direct application of domain adversarial learning to AES tasks faces significant challenges. By jointly modeling both regression and classification, we can introduce an adversarial network into the classification task to learn topic-shared features.
\subsubsection{Regression}
In addressing the need for scoring feedback in practical educational scenarios, the objective of the regression task for AES is to predict both the holistic score and the multi-trait scores for each essay. Therefore, we consider the holistic score as one of the trait scores and establish $K$ trait scorers. The initial scorer evaluates the essay's holistic score, while the remaining $K-1$ scorers independently assess the corresponding $K-1$ traits. 

Specifically, to further enhance essay scoring performance, following~\cite{ridley2020prompt}, we concatenate the essay's feature representation $\bm{h}_{\text{[CLS]}}$ with handcrafted features $\bm{f}$ as $\bm{h} = [{\bm{h}_{\text{[CLS]}}};\bm{f}]$.
% Specifically, to further enhance essay scoring performance, we first concatenate the essay's feature representation $\bm{h}_{[CLS]}$ with handcrafted features $\bm{f}$, as described in [20]. These features include include attributes such as length, readability, complexity, richness, and sentiment, thereby enriching the feature representation of the essay $x$ as:
% \begin{equation}
%     \bm{h} = [{\bm{h}_{[CLS]}};\bm{f}],
% \end{equation} 
% where $[;]$ denotes the concatenation operation. 
We then input $\bm{h}$ into a nonlinear feature transformation layer to obtain its corresponding feature representation for the $j$th scoring task: 
\begin{equation}
{\bm{h}'_j} = relu({\bm{w}_j} \cdot \bm{h} + {\bm{b}_j}) ,
\end{equation}
where ${\bm{w}_j}$  and $b_j$ are the weight matrix and the bias term of the nonlinear layer, respectively, $relu$ is the activation function. 

% For the regression task, we aim to predict the holistic and multi-trait scores of each essay. We define $K$ scorers, with the first scorer evaluates the essay holistically, while the remaining $K-1$ scorers assess the corresponding $K-1$ traits. Inspired by CTS[20], we introduce a handcrafted, topic-agnostic feature set to analyze the quality of essays from different perspectives. It contains a series of features such as length, readability, complexity, richness, and sentiment. These features are represented as a vector $\bm f$, which is concatenated with the feature representation $\bm{h}_{[CLS]}$ to enrich the feature representation of the essay $x$ as:

Generally, the holistic score of an essay is obtained by comprehensively considering the scores of various traits. Moreover, there exists an intrinsic correlation among these traits. For example, an essay that excels in word choice is likely to perform well in conventions as well. Hence, inspired by~\cite{ridley2021automated}, we apply the attention mechanism~\cite{vaswani2017attention} to trait scorers. This mechanism facilitates information interaction between each trait and the others, and can be described as follows:

\begin{equation} \bm{F} = [{\bm{h}'_1},{\bm{h}'_2},...,{\bm{h}'_K}] ,
% \end{equation}
% \begin{equation} 
\bm{v}_i^j = \frac{{\exp (score({\bm{h}'_j},{\bm{F}_{ - j,i}}))}}{{\sum\nolimits_l {\exp (score({\bm{h}'_j},\bm{F}{}_{ - j,l}))} }} ,\end{equation}
\begin{equation} {\bm{o}_j} = \sum {\bm{v}_i^j{{\bm{F}}_{ - j,i}}}  ,
% \end{equation}
% \begin{equation} 
{\bm{n}_j} = [{\bm{h}'_j};{\bm{o}_j}] ,\end{equation}
where $\bm{F}$ is the concatenation of feature representations for each trait scoring task, and $\bm{v}_i^j$ represents the attention weight of the $i$-th trait on the $j$-th trait. We then compute the trait attention feature ${\bm{o}_j}$ with the weighted summation operation on ${{F}_{ - j,i}}$ using $\bm{v}_i^j$. We concatenate ${\bm{h}'_j}$ and ${\bm{o}_j}$ to get the final feature representation ${\bm{n}_j}$.
% where $\bm{F}$ is the concatenation of feature representations for each trait scoring task. $\bm{v}_i^j$ is computed by dot product operation and softmax operation, which represents the attention weight of the $i$-th trait on the $j$-th trait. Then, we compute the trait attention feature ${\bm{o}_j}$ with the weighted summation operation on ${{F}_{ - j,i}}$ using $\bm{v}_i^j$. Both trait feature ${\bm{h}'_j}$ and trait attention feature ${\bm{o}_j}$ contain certain information, so we concatenate ${\bm{h}'_j}$ and ${\bm{o}_j}$ to get the final feature representation ${\bm{n}_j}$.

Finally, for each trait scoring task, a linear layer with a sigmoid activation function $\sigma $ maps ${\bm{n}_j}$ into a score value:
\begin{equation}
{\hat y_j} = \sigma (\bm{w}_j^y \cdot {\bm{n}_j} + b_j^y),
\end{equation}
where ${\hat y_j}$ is the predicted score for the $j$-th trait, $\bm{w}_j^y$ and $b_j^y$ are the weight matrix and the corresponding bias term, respectively. We use the mean square error ( MSE ) as a loss function for each trait scoring task. Given $N$ source topic essays and $K$ traits, the final scoring loss is computed as follows:
% \begin{equation} 
\begin{gather}
 {\mathcal{L}_{mse}} = \sum\limits_{n = 1}^N {{\mathcal{L}_n}},\\
{\mathcal{L}_n} = \frac{1}{{{\rm{|}}{S_n}{\rm{|}}\times K}}\sum\limits_{i = 1}^{|{S_n}|} {\sum\limits_{j = 1}^K {{{({{\hat y}_{i,j}} - {y_{i,j}})}^2}} }.
\end{gather}
% \end{equation}

Note that each topic may exhibit distinct traits due to varying genres (see Appendix A). Consequently, not every essay includes all trait scores. We employ the mask operation described in~\cite{ridley2021automated} to account for traits lacking true scores during the loss calculation process.

% Notably, each topic may exhibit distinct traits due to the varying genres they encompass (see details in Table ~\ref{table:ASAP++}).  For instance, topic 1 pertains to the argumentative genre, whereas topic 7 pertains to the narrative genre. Consequently, topic 7 lacks the word choice and fluency traits of topic 1 (see details in Table ~\ref{table:ASAP++}).
% This means that not every essay includes all trait scores. Therefore, we introduce a masking mechanism to account for traits that lack true scores during the loss calculation process:

% \begin{gather}
% \forall u_i \in \bm{U},{\kern 1pt} {\kern 1pt} mas{k_i} = \left\{ \begin{array}{l}
% 1,{\kern 1pt} {\kern 1pt} {\kern 1pt} {\kern 1pt} if{\kern 1pt} {u_i} \in {\bm{U}_A}\\
% {\rm{0}},{\kern 1pt} {\kern 1pt} {\kern 1pt} otherwise
% \end{array} \right., \\
%  {y_i} = {y_i} \otimes mas{k_i}{\kern 3pt},  {\hat y_i} = {\hat y_i} \otimes mas{k_i}{\kern 3pt} ,
%  \end{gather}
% where $u_i$ denotes the $i$-th trait, $\bm{U}$ denotes all possible traits, $\bm{U}_A$ denotes the traits of the current essay, and $mas{k_i} \in [0,1]$ denotes the mask of the $i$-th trait. We perform the element-wise product operation of $mas{k_i}$ with $y_i$, ${\hat y_i}$, which is $0$ and corresponds to the MSE loss of $0$ when the current essay does not have the $i$-th trait score.

\subsubsection{Classification}

% Unlike existing automated essay scoring (AES) methods that model AES solely as a classification or regression task, we jointly model the regression and classification tasks. This joint modeling approach allows for the learning of shared features between the two tasks, leading to a mutual improvement in their performance. In addition, domain adversarial training has proven highly effective in domain adaptation by learning domain-invariant feature representations through a min-max game between a domain discriminator and a feature extractor. However, as highlighted by [16], aligning the feature representations of two domains inevitably alters the feature scale. While classification tasks demonstrate robustness to variations in feature scale, regression tasks do not exhibit the same resilience. Consequently, the direct application of domain adversarial learning to AES tasks faces significant challenges. By jointly modeling both regression and classification, we can introduce an adversarial network into the classification task to learn topic-shared features.
% based on the classification task, a domain adversarial network is introduced to learn the invariant feature representation between topics, thereby enhancing the model's robustness to feature scale changes.
For classification task, we perform holistic score classification by first dividing the holistic scores into four levels: excellent, good, moderate, and poor (details in Appendix A). 
% For the holistic score classification task, we divided the holistic scores into 4 categories of excellent, good, moderate, and poor on a certain scale due to the different ranges of holistic scores for each topic (see details in Table ~\ref{table:ASAP++}).
We then input each essay $x$ into a classifier $C$ which consists of an MLP layer with a softmax activation function: $\hat c = C(\bm{h})$.
% An MLP and a softmax function are applied to the classifier:
% \begin{equation} 
% % \hat c = softmax(MLP(\bm{h})),
% \hat c = C(\bm{h}).
% \end{equation}
% where $M$ consists of an MLP layer and a softmax activation function. 
For essays from each source topic, we compute the cross-entropy loss between the predicted labels and their true labels as the classification loss: \begin{equation}
{\mathcal{L}_{c,j}} =  - \frac{1}{{|{\bm{S}_j}|}}\sum\limits_{i = 1}^{|{\bm{S}_{\rm{j}}}|} {{c_i}\log {{\hat c}_i}}.
\end{equation}
Given $N$ source topics, the holistic score classification loss is calculated as follows:

\begin{equation} 
{\mathcal{L}_{sce}} =  \sum\limits_{j = 1}^{N} {\mathcal{L}_{c,j}}.
\end{equation}

% Note that, the topic-specific prompt for each source topic can also be learned by optimizing the source topic classification loss. 
\subsection{Adversarial Training for Topic-shared Prompt Learning}

Adversarial training has proven to be effective in learning domain-invariant feature representations through a min-max game between a domain discriminator and a feature extractor. 
% The domain discriminator is trained to identify whether an essay is from the source or target domain, while the feature extractor is trained to confuse the domain discriminator. 
In this paper, we employ adversarial training in a prompt-tuning paradigm to learn the topic-shared prompt. The topic discriminator is trained to determine whether a sample comes from the source or the target topic, while the topic-shared prompt is trained to confuse the topic discriminator with the parameters of the PLM remaining fixed. This means that adversarial training occurs between the learnable topic-shared prompt and the discriminator.

% Domain adversarial network contains a feature extractor and a domain discriminator. The domain discriminator is trained to determine whether a essay comes from the source or the target domain, while the feature extractor is trained to confuse the domain discriminator. Through adversarial training of the feature extractor and the discriminator, the former learns the shared features between the source and target domain. In this paper, we introduce domain adversarial training to learn shared feature between source and target topic. It is worth noting that since the parameters of the PLM are frozen, the adversarial training is between the learnable topic-agnostic prompt and the discriminator. 

% In order to transfer the common knowledge from the source domain to the target domain and learn invariant features between different domains, we introduce adversarial training based on the essay classification task. More intuitively, ATOPT guides PLM to acquire topic invariant features by learning topic-agnostic prompt through adversarial training.

% Specifically, given $N$ source domains, we design a domain discriminator ${g_i}:{\mathbb{R}^h} \to \{ 0,1\} $ for each pair of source domain ${\bm{S}_i}$ and target domain $\bm{T}$ to discriminate whether the input essay derives from the target domain or the source domain, where 0 indicates that the prediction result is the source domain and 1 indicates that the prediction result is the target domain. We denote all $N$ domain discriminators as $\bm{G} = {\{ {g_i}\} _{1 \le i \le N}}$.
Specifically, we design a topic discriminator ${g_i}:{\mathbb{R}^h} \to \{ 0,1\} $ for each pair of source topic ${\bm{S}_i}$ and target topic $\bm{T}$ to discriminate whether the input essay derives from the target topic or the source topic. We denote all $N$ topic discriminators as $\bm{G} = {\{ {g_i}\} _{1 \le i \le N}}$. For an essay $x$ originating from the source topic $i$ or the target topic, we input it into its corresponding discriminator to obtain the probability of belonging to the target topic by:

\begin{equation}
p(y = 1|x) = \frac{{\exp (g_{i,1}({\bm{h}_{[CLS]}}))}}{{\sum\nolimits_{j \in \{ 0,1\} } {\exp (g_{i,j}({\bm{h}_{[CLS]}}))} }}.
\end{equation}

Given $N$ source domain datasets $\bm{S} = \{ {\bm{S}_i}\} _{i = 1}^N$ and target domain dataset $\bm T$, we can compute the cross entropy loss between the predicted and true values as follows:
% the training objective of the discriminatoris to minimize the cross-entropy loss between the predictedand true values:

 \begin{equation} \begin{split}
 % {L_{adv}}(\bm{S},\bm{T},\bm{G}) =  &- \sum\limits_{i = 1}^N {[\sum\limits_{x \in {\bm{S}_i}} {\log p(y = 0|x)} } \\
 % &+ \sum\limits_{x' \in \bm{T}} {\log (1 - p(y = 0|x'))}
 % \end{split} .
 {\mathcal{L}_{adv}} =  - \sum\limits_{i = 1}^N {[\sum\limits_{x \in {\bm{S}_i}} {\log(1- p(y = 1|x))} } +
 \sum\limits_{x' \in \bm{T}} {\log  p(y = 1|x')]}
\end{split} .
\end{equation}

% \begin{equation} \begin{split}
% % {L_{adv}}(\bm{S},\bm{T},\bm{G}) =  &- \sum\limits_{i = 1}^N {[\sum\limits_{x \in {\bm{S}_i}} {\log p(y = 0|x)} } \\
% % &+ \sum\limits_{x' \in \bm{T}} {\log (1 - p(y = 0|x'))}
% % \end{split} .
% {\mathcal{L}_{adv}} =  - \sum\limits_{i = 1}^N {[\sum\limits_{x \in {\bm{S}_i}} {\log(1- p(y = 1|x))} } +\\
% \sum\limits_{x' \in \bm{T}} {\log  p(y = 1|x')]}
% \end{split} .
% \end{equation}

Following ~\cite{ganin2015unsupervised}, we apply a Gradient Reversal Layer (GRL) to the discriminator, reversing its gradient direction during training.

% In this paper, we apply a simple and efficient Gradient Reversal Layer \cite{ganin2015unsupervised} to the discriminator, which makes the network training goal reversed before and after the discriminator by multiplying the gradient of the discriminator with a time-varying parameter. Intuitively, the goal of the discriminator is still to minimize the discriminative loss, but the training goal of the feature extractor is transformed into maximizing the discriminative loss.
\begin{algorithm}[t]\small
	\caption{The training process of ATOP.}
	\label{algorithm:training process} 
	
	\begin{algorithmic}[1]
		\renewcommand{\algorithmicrequire}{ \textbf{Input:}} %Use Input in the format of Algorithm
		\REQUIRE ~~\\
		\renewcommand{\arraystretch}{1}
	{Training samples of $N$ source topic datasets $\bm{S_{}} = \{ ({\bm{X}_i},{\bm{Y}_i^{}, \bm{C}_i^{})\}} _{i = 1}^N$ and a target topic dataset $\bm{T} = \{ \bm{x}_t^{}\} _{t = 1}^{|\bm {T}|}$.}%输入参数
    
            \renewcommand{\algorithmicrequire}{ \textbf{Input parameters:}}
		\REQUIRE ~~\\
            {PLM ${\theta_M}$, topic-shared prompt embedding ${\theta_{A}}={\bm{p}}_{shared}$, topic-specific prompt embedding ${\theta_{S}}= \{{\bm p}_{specific}^i\} _{i = 1}^{N+1}$; classifiers ${\theta _{C}}$; discriminators ${\theta _G}$; trait score predictors ${\theta _T} $;
            \renewcommand{\algorithmicrequire}{\textbf{Hyperparameters:}} 
            \REQUIRE ~~\\
            learning rate $\eta$; trade-off parameters $\alpha$ and $\beta$.}%输入参数            
		\renewcommand{\algorithmicrequire}{ \textbf{Output:}}
		\REQUIRE ~~\\
		Trained parameters ${\theta _{A,S,C,G,T,M}}$.
        
            \renewcommand{\algorithmicrequire}{ \textbf{Initalization:}}
            \REQUIRE ~~\\
            Memory bank $\bm{M}$ for target prompt.
		
		\FOR{each mini-batch}
		\STATE Minibatch data of source topics and target topic $x = \{ {x_s},{x_t}\}$
        \STATE \textcolor[rgb]{0.8,0.549,0}{/* Training topic-shared prompt. */}
        \STATE Frozen topic-specific prompts ${\theta_{S}}$ 
        \STATE ${\mathcal{L}_{ce}} \leftarrow {\mathcal{L}_{ce}}(x;{\theta _{A,C,M}})$
        \STATE${\mathcal{L}_{mse}} \leftarrow {\mathcal{L}_{mse}}(x;{\theta _{A,T,M}})$
        \STATE${\mathcal{L}_{adv}} \leftarrow {\mathcal{L}_{adv}}(x;{\theta _{A,G,M}})$
        \STATE\textcolor[rgb]{0.8,0.549,0}{/* Update parameters. */}
        \STATE${\theta _{C,T,G,A}} \leftarrow {\theta _{C,T,G,A}} - {\nabla _{{\theta _{C,T,G,A}}}}{\mathcal{L}_{total}}$
        \STATE\textcolor[rgb]{0.8,0.549,0}{/* Training topic-specific prompt. */}
        \STATE Frozen topic-shared prompt ${\theta_{A}}$ 
        \STATE${\mathcal{L}_{ce}} \leftarrow {\mathcal{L}_{ce}}(x;{\theta _{S,C,M}})$
        \STATE${\mathcal{L}_{mse}} \leftarrow {\mathcal{L}_{mse}}(x;{\theta _{S,T,M}})$
        \STATE\textcolor[rgb]{0.8,0.549,0}{/* Update parameters. */}
        \STATE${\theta _{C,T,S}} \leftarrow {\theta _{C,T,S}} - {\nabla _{{\theta _{C,T,S}}}}({\mathcal{L}_{ce}}+{\alpha \mathcal{L}_{{\rm{mse}}}}) $
	\ENDFOR
	\end{algorithmic}
\end{algorithm}
\renewcommand{\thealgorithm}{\arabic{algorithm}} % 仅影响算法编号
\renewcommand{\thetable}{\arabic{table}} % 确保表格编号是阿拉伯数字
\subsection{Pseudo-labeling for Topic-specific Prompt Learning}
% We design the topic-specific soft prompts for each topic to learn topic-related information. However, the target topic soft prompt can not be learned due to the invisibility of target topic's labels. To solve this problem, we introduce a pseudo-labeling method for target domain data. Additionally, to distinguish between topic-agnostic and topic-specific soft prompts, we freeze the other prompt while training one of them. We will explain the specific training process in detail in subsection 3.5.
In this paper, we utilize topic-specific prompt to extract topic-specific knowledge from PLMs. For each source topic, we learn the corresponding topic-specific prompt by optimizing its classification loss. For the target topic, which lacks score labels, we initially assign pseudo-labels and then optimize the classification loss based on these pseudo-labels to learn the corresponding topic-specific prompt.

We employ a straightforward yet effective pseudo-labeling method that leverages the local data structure stored in a memory bank to generate pseudo-labels for target essays~\cite{liang2021domain}. Specifically, for each target essay ${{x}_t}$ in a mini-batch, we first obtain its feature representation $\bm{f_t}=\bm{h}_t$ and the corresponding soft label $\bm{p}_t=C(\bm{h}_t)$. To mitigate noise interference during label prediction, we apply sharpening operations to the soft labels:
\begin{equation}
\tilde p_{t,k}^{} = p_{t,k}^\tau /\sum\limits_k {p_{t,k}^\tau },
\end{equation}
where $\tau $ is a temperature coefficient, and ${p_{t,k}}$ denotes the prediction probability for category $k$. Features and soft labels of all target topic essays are stored in a memory bank denoted as ${B}$, which is updated using an exponential moving averaging strategy in each iteration:
% We utilize a simple but effective pseudo-labeling method based on Neighborhood Aggregation (NA) \cite{liang2021domain}, which consists of a memory bank storing features and soft labels of all target domain data $B$ and generating pseudo label by aggregating the neighbor information of the current sample. Specifically, we first randomly initialize the memory bank ${B}$ and update $B$ with the exponential moving averaging strategy:
\begin{equation} 
{\overline {\bm{f}}_t} = \lambda {\overline {\bm{f}}_t} + (1 - \lambda ){\bm{f}_t} , \  
{\overline{\bm{p}}_t} = \lambda {\overline {\bm{p}}_t} + (1 - \lambda ){\bm{p}_t},
\end{equation}
where $\lambda $ is a smoothing parameter. 
% ${\tilde {\bm{f}_t}}$ and ${\tilde{\bm{p}_t}}$ correspond to the feature representation and soft label of the current sample at the time step $t$ in the memory bank. Here, ${\bm{f}_t}$ is the feature representation $\bm{h}$ of an essay $x$, $\bm{p}_t^{} = C({\bm{f}_t})$ is the predicted soft label by the classifier $C$.

Then, we identify the $m$ nearest-neighbor essays of $x_t$ by computing the cosine similarity between its feature vector ${\bm{h}_t}$ and those of all essays stored in the memory bank $B$. 
% Unlike prior methods that are restricted to mini-batches, our memory bank enables us to search for informative neighbors across the entire set of target essays. 
Next, we aggregate the soft labels from the $m$ neighbor essays to generate a soft label for $x_t$. The final pseudo-label is determined by selecting the category with the highest probability in the soft labeling:
\begin{equation}
{\hat y_t} = \mathop {\arg \max }\limits_k {\hat p_{t,k}},\ 
\hat{\bm{p}}_t = \frac{1}{m}\sum\limits_{i \in {\bm{N}_t}} {\overline{\bm{p}}_t^i}, 
\end{equation}
where $\bm{N}_t$ represents the index set of the $m$ nearest neighbors of essay ${x}_t$. Based on the generated pseudo-labels, we compute the cross-entropy loss for the target topic:
\begin{equation}
    {\mathcal{L}_{tce}} =  - \frac{1}{{|\bm{T}|}}\sum\limits_{t = 1}^{|\bm{T}|} {\sum\limits_{k = 0}^{{\rm{C}} - 1} {{\bm{\hat y}_{t,k}}\log {C_k}({\bm{h}_t})} } .
\end{equation}
% To avoid noise interference in label prediction, we perform sharpening operations on soft labels:
% \begin{equation}\tilde p_{t,k}^{} = p_{t,k}^\tau /\sum\limits_{t = 1}^{|\bm{T}|} {p_{t,k}^\tau }  ,\end{equation}
% where $\tau $ is the temperature coefficient, and ${p_{t,k}}$ represents the prediction probability of sample ${x}_t$ on category $k$. 
% Considering that low confidence pseudo labels may interfere with the learning process of the model, we directly regard the corresponding highest prediction probability as the confidence value and add it to the cross-entropy loss:
% \begin{equation}
% {L_{tce}} =  - \frac{1}{{|\bm{T}|}}\sum\limits_{t = 1}^{|\bm{T}|} {\sum\limits_{k = 0}^{\text{C} - 1} {{\bm{w}_t} * } } {\tilde y_{t,k}}\log {C_k}({\bm{f}_t}) ,
% \end{equation}

In summary, the holistic score classification loss for all topics is as follows:
\begin{equation}{\mathcal{L}_{ce}} = \mathcal{L}_{sce} + \mathcal{L}_{tce}  .\end{equation}

By optimizing this holistic classification loss, we can learn a topic-specific prompt for each topic.
\begin{table*}[htbp]\centering
\caption{Average QWK of all traits on each topic.
% Comparison of regression results of all methods for the topic dimension. The QWK value for each topic is calculated as the average of the holistic score and the QWK values of all trait scores. 
% {\dagger} \textrm{ refers to the results from PMAES~\cite{chen2023pmaes}.}
}
 \begin{tabular*}{\hsize}{@{}@{\extracolsep{\fill}}lccccccccccl@{}}
\toprule
\multicolumn{2}{l}{Methods}              & Topic 1         & Topic 2         & Topic 3         & Topic 4         & Topic 5         & Topic 6         & Topic 7         & Topic 8        & \bf{Avg}              \\
\midrule

\multicolumn{2}{l}{Hi-att 
% \dagger
}       & 0.315           & 0.478          & 0.317          & 0.478         & 0.375         & 0.357          & 0.205         & 0.265                 & 0.349         \\

\multicolumn{2}{l}{AES aug 
% \dagger
}       & 0.330         & 0.518        & 0.299        & 0.477          & 0.341         & 0.399         & 0.162         & 0.200            &0.341        \\
\multicolumn{2}{l}{Bert-FT}       & 0.542         & 0.546        & 0.574        & 0.603          & 0.630         & 0.459         & 0.256         & 0.235            & 0.481        \\
\multicolumn{2}{l}{PT-V2}       & 0.570         & 0.562        & \textbf{0.623}        & \textbf{0.655}          & \uline{0.668}         & 0.506         & 0.346         & 0.409            & 0.542        \\
\multicolumn{2}{l}{PAES 
% \dagger
}     & 0.605            & 0.522         & 0.575         & 0.606          & 0.634          & 0.545          & 0.356         &  0.447                  & 0.536          \\

\multicolumn{2}{l}{CTS 
% \dagger
}      & 0.623          &  0.540         & 0.592         &{0.623}         & 0.613         &  0.548        &  0.384         & 0.504                &  0.553      \\

\multicolumn{2}{l}{PMAES 
% \dagger
}      & \textbf{0.656}          &  {0.553}         & {0.598}         &{0.606}         & {0.626}         &  {0.572}        &  {0.386}         & {0.530}            & {0.566}       \\
\multicolumn{2}{l}{PLAES 
% \dagger
}      & \uline{0.648}          &  \uline{0.563}         & 0.604        &0.623         & 0.634         &  \uline{0.593}        &  \uline{0.403}         & \uline{0.533}            & \uline{0.575}       \\

\multicolumn{2}{l}{ATOP (Ours)}    &{0.642}             & \textbf{0.583}    & \uline{0.620} & \uline{0.637} & \textbf{0.675}    & \textbf{0.594} & \textbf{0.436}    & \textbf{0.565}       & \textbf{0.594}   \\
\bottomrule
% \end{tabular}
\end{tabular*}
% }
\label{table:compare:topic}
\end{table*}

\begin{table*}[htbp]\centering
\caption{Average QWK for each trait over all topics.
% Comparison of regression results of all methods for each trait. The QWK value for each trait is calculated as the average of the QWK values of all topics.
% \dagger \textrm{ refers to the results from PMAES~\cite{chen2023pmaes}.}
}
\begin{tabular*}{\hsize}{@{}@{\extracolsep{\fill}}lccccccccccccc@{}}
\toprule
\multicolumn{2}{l}{Methods}              & Holistic         & Cont         & Org         & WC         & SF         & Conv         & TA         & Lan   & Nar      & \bf{Avg}             \\
\midrule

\multicolumn{2}{l}{Hi-att 
% \dagger
}       & 0.453           & 0.348          & 0.243          & 0.416         & 0.428         & 0.244          & 0.309         & 0.293        & 0.379         & 0.346         \\

\multicolumn{2}{l}{AES aug 
% \dagger
}       & 0.402         & 0.342        & 0.256        & 0.402          & 0.432         & 0.239         & 0.331         & 0.313         & 0.377        &0.344        \\

\multicolumn{2}{l}{Bert-FT}       & 0.499         & 0.492        & 0.370        & 0.473          & 0.408         & 0.331         & { 0.591}         & 0.529            & 0.608  & 0.478      \\
\multicolumn{2}{l}{PT-V2}      &0.622 & 0.521         & 0.393        & \uline{0.601}        & \textbf{0.596}          & 0.336         & 0.579         & \textbf{0.598}         & 0.625            & 0.542        \\
\multicolumn{2}{l}{PAES 
% \dagger
}     & 0.657            & 0.539         & 0.414         & 0.531          & 0.536          & 0.357          & 0.570         &  0.531        & 0.605          & 0.527          \\

\multicolumn{2}{l}{CTS 
% \dagger
}      & 0.670          &  0.555         & 0.458         &0.557         & 0.545         &  0.412        &  0.565         & 0.536         &  0.608        &  0.545      \\

 \multicolumn{2}{l}{PMAES }      & {0.671}          &  {0.567}         & {0.481}         &0.584         & 0.582         &  {0.421}        &  {0.584}         & {0.545}         &  {0.614}        & { 0.561}       \\

\multicolumn{2}{l}{PLAES 
% \dagger
}      & \uline{0.673}          &  \uline{0.574}         & \uline{0.491}         &{0.579}         & {0.580}         &  \uline{0.447}        &  \uline{0.601}         & 0.554         &  \uline{0.631}        & \uline{0.570}       \\

\multicolumn{2}{l}{ATOP (Ours)}    & \textbf{0.683}             & \textbf{0.592}    & \textbf{0.531} & \textbf{0.610} & \uline{0.591}    & \textbf{0.456} & \textbf{0.618}    & \uline{0.583}    & \textbf{0.644}    & \textbf{0.590}   \\
\bottomrule
% \end{tabular}
\end{tabular*}
\label{table:compare:trait}
\end{table*}

\subsection{Learning Procedure}

In this paper, our ATOP model frames AES as a joint regression and classification task within the prompt-tuning paradigm, optimizing topic-shared and topic-specific prompts to induce relevant knowledge in a PLM. The model utilizes three types of loss functions: regression loss, classification loss, and adversarial loss, with the overall loss defined as: 
\begin{equation} {\mathcal{L}_{total}} = {\mathcal{L}_{ce}} + \alpha {\mathcal{L}_{mse}} + \beta {\mathcal{L}_{adv}} ,\end{equation}
where $\alpha$ and $\beta$ are trade-off parameters that balance the weights of the different losses. For clarity, the entire training process of ATOP is summarized in Algorithm~\ref{algorithm:training process}.

% In each iteration, the given mini batch data for each source and target domain are used to train soft prompts, $N$ discriminators, classifiers, and $K$ trait raters. First, we freeze the topic-specific soft prompt parameters. The classification loss, trait loss, and adversarial loss are computed in lines 5-7, which are used to update the parameters of the corresponding modules and the topic-shared soft prompt (line 9). Then, to train topic-specific soft prompt, we freeze topic-shared soft prompt and discriminators. The classification loss and the regression trait loss are computed in lines 12-13, which are used to update the unfrozen parameters (line 15).

In each iteration, mini-batch data from both the source and target domains are used to train the soft prompts, 
$N$ discriminators, classifiers, and $K$ trait raters. First, the topic-specific soft prompt parameters are frozen. The classification loss, trait loss, and adversarial loss are computed (lines 5–7) and used to update the corresponding modules along with the topic-shared soft prompt (line 9). Next, to train the topic-specific soft prompt, the topic-shared soft prompt and discriminators are frozen. The classification loss and regression trait loss are then computed (lines 12–13) and used to update the remaining trainable parameters (line 15).

\section{Experiments}
% \subsection{Dataset and Evaluation Metric}

% We evaluated our model using the ASAP++ dataset~\cite{mathias2018asap++}, an extended version of the ASAP dataset\footnote{https://www.kaggle.com/c/asap-aes/data} from the Automated Student Assessment Prize competition, which builds on the original ASAP dataset with trait scores for all topics. The ASAP++ dataset comprises 12,987 essays across eight different topics and encompasses three genres: narrative, argumentative, and source-dependent. In the classification task, we divided essay scores into four categories. Specifically, the essay scores are first normalized to the range [0, 1]. Then we assign scores in the intervals (0, 0.4), (0.4, 0.6), (0.6, 0.8), and (0.8, 1) to their corresponding grades (excellent, good, moderate, and poor), respectively. Meanwhile, some traits only exist in separate topics, and the model could not learn relevant information from the source topic when these topics serve as target topics, so we removed these traits, such as the style trait of prompt 7 and the sound trait of prompt 8. TABLE I displays the statistics for ASAP++ datasets.

% We employed the quadratic weighted kappa (QWK) as our evaluation metric, which is widely used in AES tasks to measure the consistency between automated and human raters. 
\subsection{Dataset and Evaluation Metric}

We evaluated our model using the ASAP++ dataset~\cite{mathias2018asap++}, an extended version of the ASAP dataset\footnote{https://www.kaggle.com/c/asap-aes/data} from the Automated Student Assessment Prize competition. The ASAP++ dataset includes 12,987 essays across 8 topics, spanning three genres: narrative, argumentative, and source-dependent. Each genre involves different scoring traits. For the holistic score classification task, the essays were categorized into four levels: excellent, good, moderate, and poor—based on their scores. Detailed statistics and division criteria are provided in Appendix A. We employed the Quadratic Weighted Kappa (QWK) as our evaluation metric, which is widely used in AES tasks to measure the consistency between automated and human raters.

\subsection{Implementation Details}
% For all of our experiments, we use BERT-base model as the backbone. The output feature dimension of PLM is 768 dimensions, which is reduced to 100 dimensions by a linear layer. During training, the model is trained for 30 epochs with batch size of 4 for each domain. The loss trade-off parameters $\alpha$ and $\beta$ are 10 and 1, respectively. The prompt lengths $m$ and $n$ are both 8. We optimized the model parameters using the Adam optimizer with an initialized learning rate of 0.01, a decay rate of 0.9, and a decay step of 2000. For the multi-trait AES task, we treat all traits as equally important and take the epoch with the highest average QWK for all traits as the best result. Our model is implemented in python using pytorch and train on 4 Nvidia Gefore GTX TITANs.

In our experimental implementation, we utilized the Bert-base model as the PLM. During training, we designated one of the topics as the target topic and the remaining seven topics as source topics to train the model. The trade-off parameters $\alpha$ and $\beta$ were set to 10 and 1, respectively, and the lengths $m$ and $n$ of the topic-agnostic and topic-specific prompts were both set to 8. Detailed implementation procedures and parameter settings are provided in Appendix B.

\subsection{Baselines}

% To evaluate the effectiveness of our proposed model, we compared it against eight state-of-the-art methods across three tasks: holistic score classification, holistic score regression, and multi-trait score regression. Baseline models include Hi-att~\cite{dong2017attention}, AES aug~\cite{hussein2020trait}, Bert-FT~\cite{devlin2019bert}, PAES~\cite{ridley2020prompt}, CTS~\cite{ridley2021automated}, PMAES~\cite{chen2023pmaes}, PLAES~\cite{li2024plaes}, and LLM-fs~\cite{stahl2024exploring}. Note that since LLM-fs performs only holistic score regression for essays, we report its experimental results solely for the overall regression task.
To evaluate the effectiveness of our proposed model, we compared it against nine state-of-the-art methods across three tasks: holistic score classification, holistic score regression, and multi-trait score regression. Baseline models include Hi-att~\cite{dong2017attention}, AES aug~\cite{hussein2020trait}, Bert-FT~\cite{devlin2019bert}, PT-V2~\cite{liu2021p}, PAES~\cite{ridley2020prompt}, CTS~\cite{ridley2021automated}, PMAES~\cite{chen2023pmaes}, PLAES~\cite{li2024plaes} and LLM-fs~\cite{stahl2024exploring}. Note that since LLM-fs performs only holistic score regression for essays, we report its experimental results solely for the overall regression task in Appendix D.

\subsection{Results and Analysis}

% For the holistic score classification task, Table~\ref{table:classfication} shows that our method achieves the best result and outperforms the optimal baseline model by 2.7$\%$ in terms of average QWK on all topics, which demonstrates the effectiveness and advancement of our method on this task. Our experimental results from Table~\ref{table:classfication} show the poor performance of the Hi-att model, which indicates that there are significant domain differences between the source and target topics, and that directly transferring the model trained on the source topic data to the target topic leads to a significant performance degradation. In addition, three improved methods (PAES, CTS and PMAES) based on the Hi-att model have achieved better classification results, which shows that adding generic features of essays (e.g., POS features and handcrafted features) can improve the generalization of the model.
Tables~\ref{table:compare:topic} and~\ref{table:compare:trait} present the results of the multi-trait score regression task from the perspectives of topics and traits, respectively. Note that bold numbers in tables indicate the best results in each column, while underlined numbers represent the second-best. As shown in Table~\ref{table:compare:topic}, our method ATOP improves the average QWK score across different topics by 1.9\% compared to the best baseline, indicating enhanced multi-trait scoring performance in the topic dimension. Table~\ref{table:compare:trait} further demonstrates that ATOP consistently outperforms other methods across nearly all traits, particularly on topic-specific traits such as word choice and topic adherence, with QWK gains ranging from 0.9\% to 4.0\%. These results highlight the effectiveness of jointly learning topic-shared and topic-specific prompts, supported by adversarial training and pseudo-labeling, in improving model performance.

\begin{table*}[t]\centering
\caption{Ablation study of the key components on the topic dimension.} 
\begin{tabular*}{\hsize}{@{}@{\extracolsep{\fill}}lccccccccccc@{}}
\toprule
\multicolumn{2}{l}{Methods}              & Topic 1         & Topic 2         & Topic 3         & Topic 4         & Topic 5         & Topic 6         & Topic 7         & Topic 8        & \textbf{Avg}              \\
\midrule

\multicolumn{2}{l}{ATOP}    & \textbf{0.642}            & 0.583    & \textbf{0.620} &0.637 & \textbf{0.675}    & 0.594 & \textbf{0.436}    & \textbf{0.565}       & \textbf{0.594}   \\

\multicolumn{2}{l}{ w/o ${\mathcal{L}_{adv}}$ }       & 0.603           & \textbf{0.592}          & 0.605          & \textbf{0.644}         & 0.650         & \textbf{0.601}          & 0.393         & 0.511                 & 0.575         \\

\multicolumn{2}{l}{w/o $\bm{\theta_{S}}$}       & 0.634         & 0.559        & 0.618        & 0.633          & 0.671         & 0.574         & 0.431         & 0.550     &0.584        \\

\multicolumn{2}{l}{w/o ${\mathcal{L}_{ce}}$}       & 0.631         & 0.560        & 0.612        & 0.636          & 0.656         & 0.577         & 0.429         & 0.541     &0.580        \\
\bottomrule
\end{tabular*}
% }
\label{table:ablation:topic}
\end{table*}
\begin{table*}[t]\centering
\caption{Ablation study of the key components on the trait dimension.} 
\begin{tabular*}{\hsize}{@{}@{\extracolsep{\fill}}lccccccccccccc@{}}
\toprule
\multicolumn{2}{l}{Methods}              & Holistic         & Cont         & Org         & WC         & SF         & Conv         & TA         & Lan   & Nar      & \textbf{Avg}              \\
\midrule

\multicolumn{2}{l}{ATOP}    & \textbf{0.683}             & \textbf{0.592}    & \textbf{0.531} & \textbf{0.610} & \textbf{0.591}    & \textbf{0.456} & {0.618}    & {0.583}    & {0.644}    & \textbf{0.590}   \\

\multicolumn{2}{l}{ w/o ${\mathcal{L}_{adv}}$ }       & 0.639           & 0.579          & 0.485          & 0.562         & {0.586}         & 0.431          & \textbf{0.625}         & \textbf{0.591}        & \textbf{0.647}         & 0.572    \\

\multicolumn{2}{l}{w/o $\bm{\theta_{S}}$}       & 0.664         & 0.589        & 0.525        & 0.599          & 0.576         & 0.438         & 0.612         & 0.578         & 0.630        &0.579        \\
\multicolumn{2}{l}{w/o ${\mathcal{L}_{ce}}$}       & 0.660         & 0.582        & 0.522        & 0.594          & 0.573         & 0.437         & {0.615}         & 0.573     &0.626   &0.576     \\
\bottomrule
\end{tabular*}
% }
\label{table:ablation:trait}
\end{table*}

% \begin{figure*}[t]
% \centering
% \subfloat[]{
% \includegraphics[width=0.47\linewidth]{figure/T1.pdf}
% }
% \subfloat[]{
% \includegraphics[width=0.47\linewidth]{figure/T3.pdf}
% }
% \\
% \subfloat[]{
% \includegraphics[width=0.47\linewidth]{figure/T5.pdf}
% }
% \subfloat[]{
% \includegraphics[width=0.47\linewidth]{figure/T8.pdf}
% }
% \caption{Visualization results for topic 1 (a), topic 3 (b), topic 5 (c) and topic 8 (d). Red represents the features of the source topic, green represents the features of the target topic.}
% \label{fig:visual}
% \end{figure*}

% Additionally, Table~\ref{table:compare:topic} and ~\ref{table:compare:trait} reveal that Hi-att and AES-aug produce the weakest results, indicating that directly applying models trained on source topics to target topics causes severe performance degradation due to domain drift. However, models such as PAES and CTS, despite being built upon the Hi-att backbone, demonstrate promising results, suggesting that incorporating POS features and manually crafted topic-shared features can improve model performance. Furthermore, PMAES and PLAES achieve even greater improvements over CTS, underscoring the importance of domain adaptation techniques in cross-topic automated essay scoring.

Notably, while both ATOP and Bert-FT use the Bert-base model as their backbone, ATOP significantly outperforms Bert-FT. This performance gap can be attributed to the inherent challenges of cross-topic AES. Despite fine-tuning the BERT model with all available source-topic data, Bert-FT struggles to mitigate domain drift between source and target topics, resulting in suboptimal performance. Furthermore, while PT-V2 leverages topic-shared prompts and performs well in several topics, it lacks the ability to incorporate topic-specific information, limiting its effectiveness. In contrast, ATOP jointly learns topic-shared and topic-specific prompts, enabling it to more effectively transfer and adapt relevant knowledge from BERT to the target topic.

However, our method underperforms compared to PMAES and PLAES on topic 1 in the multi-trait score regression task. This may be due to the nature of topic 1, which consists of argumentative essays that typically follow a more rigid writing structure. Unlike our method, PMAES and PLAES use POS features rather than raw word sequences, allowing them to better capture syntactic patterns, leading to superior performance on this specific topic. 

Additionally, to highlight ATOP's advantages across different tasks, we provide comparative experimental results for the holistic score regression and classification tasks in Appendix D.

% To further illustrate ATOP’s advantages across various tasks, we present comparative experimental results for holistic score classification and regression in TABLE~\ref{table:classfication} and TABLE~\ref{table:regression}. As demonstrated, our model outperforms the best baseline models in terms of average QWK values across all topics, achieving improvements of 3.4\% and 3.2\%, respectively. These results highlight the robustness and effectiveness of our model in both holistic score classification and regression tasks. Notably, as shown in TABLE~\ref{table:regression}, the performance of LLM-fs is significantly lower than that of existing cross-topic AES approaches. This underperformance can be primarily attributed to the inherent complexity of AES, coupled with the limitations of LLM-fs's complete reliance on manually designed text prompt templates to elicit LLMs' scoring ability. This disparity suggests that effectively applying LLMs to complex tasks such as AES necessitates the development of more sophisticated techniques and the integration of task-specific information to achieve competitive performance.
\begin{figure*}[htb]
\centering
\subfloat[]{
\includegraphics[width=0.3\linewidth]{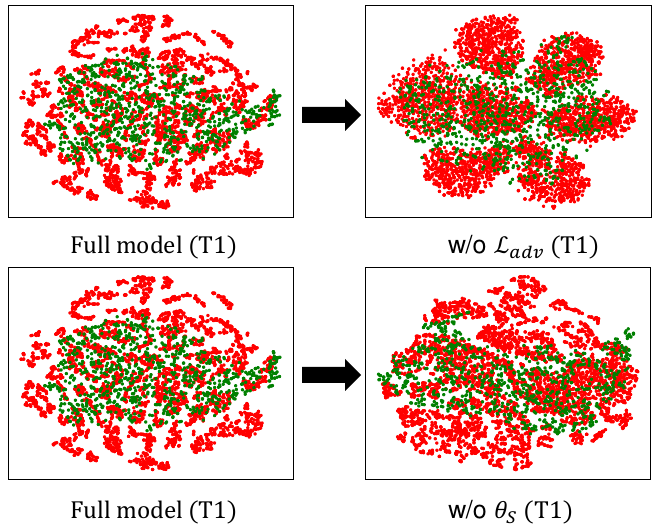}
}
\subfloat[]{
\includegraphics[width=0.3\linewidth]{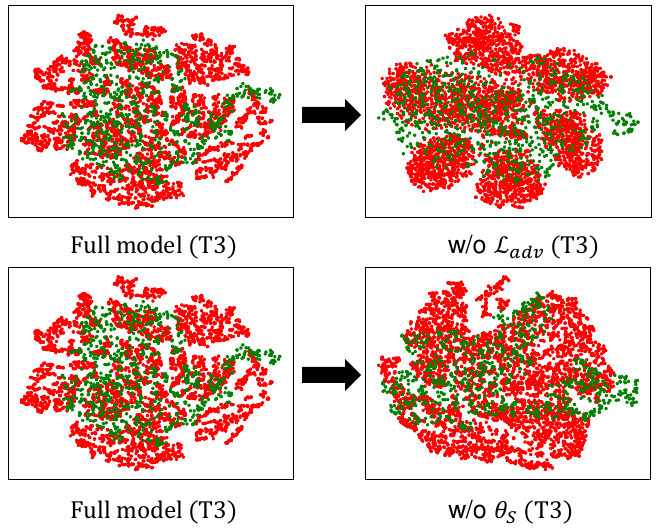}
}
\subfloat[]{
\includegraphics[width=0.3\linewidth]{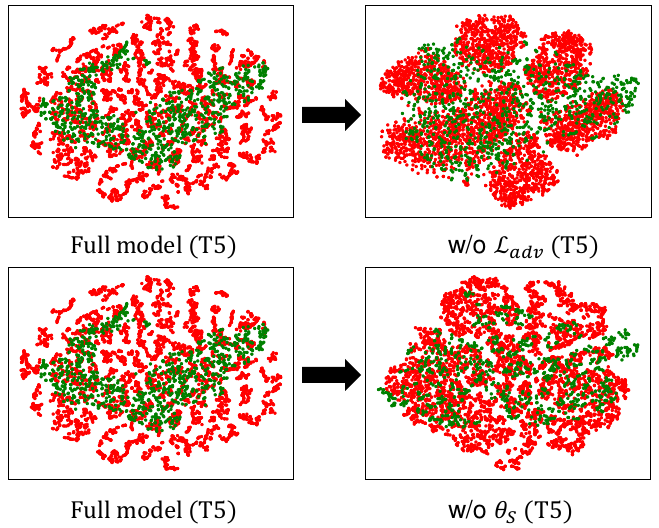}
}
% \subfloat[]{
% \includegraphics[width=0.47\linewidth]{figure/T8.pdf}
% }
\caption{Visualization of features for Topics 1, 3 and 5. Red represents the features of the source topic, green represents the features of the target topic.}
\label{fig:visual}
\end{figure*}

\subsection{Ablation Studies}

To further demonstrate the effectiveness of the key components of our method, we conducted ablation experiments from both topic and trait perspectives, as shown in Table~\ref{table:ablation:topic} and Table~\ref{table:ablation:trait}. 

In the topic dimension (Table~\ref{table:ablation:topic}), removing adversarial training, the topic-specific prompt, and classification modeling led to average performance decreases of 1.9\%, 1.0\%, and 1.4\%, respectively, which shows the effectiveness of these components. Interestingly, removing adversarial training slightly improved performance for topics 2, 4, and 6. These topics, designed for 10th-grade students, differ significantly from others, and direct alignment with other topics may result in performance degradation. Furthermore, removing the topic-specific prompt and classification modeling led to varying degrees of performance degradation for each topic, which illustrates the important role of topic-specific information and classification tasks for model training.

In the trait dimension (Table \ref{table:ablation:trait}), removing adversarial training reduced average model performance by 1.8\%. Notably, traits with low topic relevance (e.g., organization, word choice, conventions) showed more significant decreases, while traits with high topic relevance (e.g., topic adherence, language, and narrativity) improved, suggesting that adversarial training helps in learning topic-shared features. Furthermore, removing topic-specific prompts and classification modeling reduced the performance by 1.1\% and 1.4\%, respectively, further confirming the effectiveness of these components. 
% It should be noted that the classification task does not significantly improve the performance of the model, but the adversarial training module is based on the implementation of the classification task, so it is an integral part.

\subsection{Visualization Analysis}

To visually demonstrate the role of adversarial training and the topic-specific prompt, we present the feature representations of ATOP, ATOP w/o ${\mathcal{L}_{adv}}$ and ATOP w/o $\bm{\theta_{S}}$, respectively. As shown in Figure~\ref{fig:visual}, we used t-SNE~\cite{donahue2014decaf} to project the feature representations of Topics 1, 3 and 5, from both source and target topics, into a two-dimensional space. 

As shown in Figure~\ref{fig:visual}, the feature distributions of the source and target topics in the full model are relatively compact and exhibit a high degree of overlap. However, when the adversarial training module is removed, the feature representations of different source topic samples form distinct clusters. This indicates that adversarial training encourages the extraction of topic-agnostic features, as it reduces the model's ability to distinguish features by topic. Additionally, without adversarial training module, the feature distributions of the source and target topics become more separated, reflecting a reduction in cross-topic consistency. Similarly, removing the topic-specific prompts leads to unconstrained learning of the target topic's feature distribution, resulting in scattered and disorganized representations. This highlights the crucial role of topic-specific prompts in guiding the model to learn well-structured and compact topic-specific feature representations, thereby preserving the coherence of the feature space.

Notably, Figure 2 (c) illustrates that when Topic 5 is designated as the target topic, the full model not only learns a compact and regular feature distribution for Topic 5 but also aligns it effectively with the source topic distributions. This alignment corresponds to the model's best performance on Topic 5, as reported in Table 1, further emphasizing the effectiveness of combining adversarial training with topic-specific prompts in enhancing cross-topic feature learning.

\subsection{Hyperparameter Analysis}\label{sec:hyper}
\begin{figure}[ht]
\centering
\subfloat[]{
\includegraphics[width=0.45\linewidth]{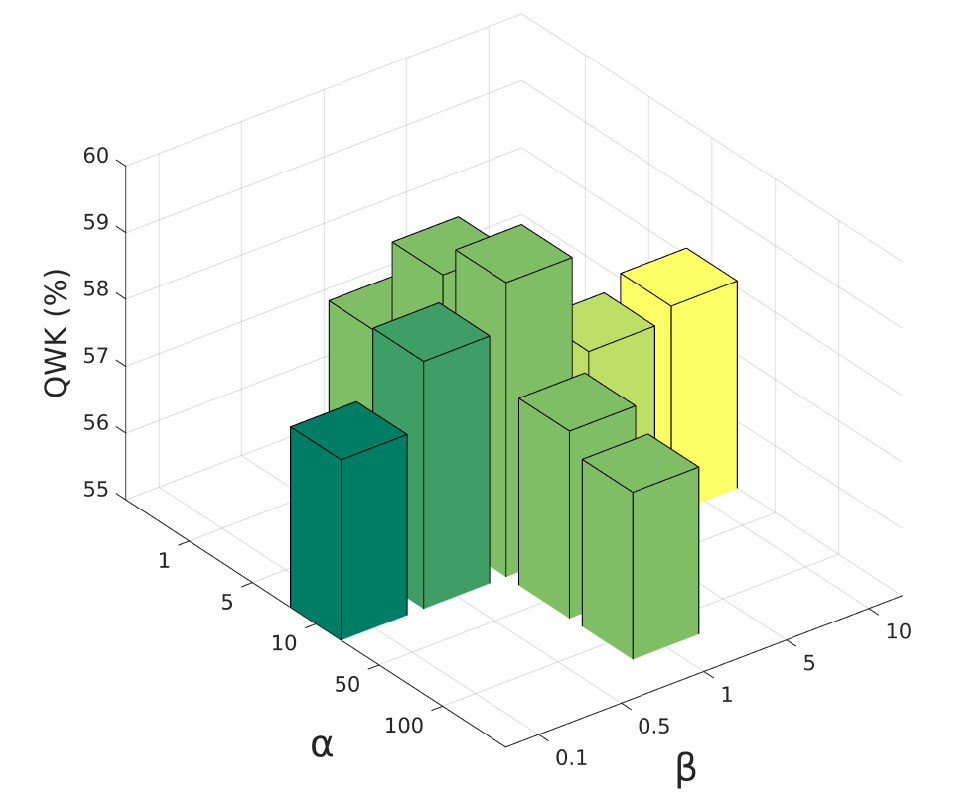}
}
\subfloat[]{
\includegraphics[width=0.45\linewidth]{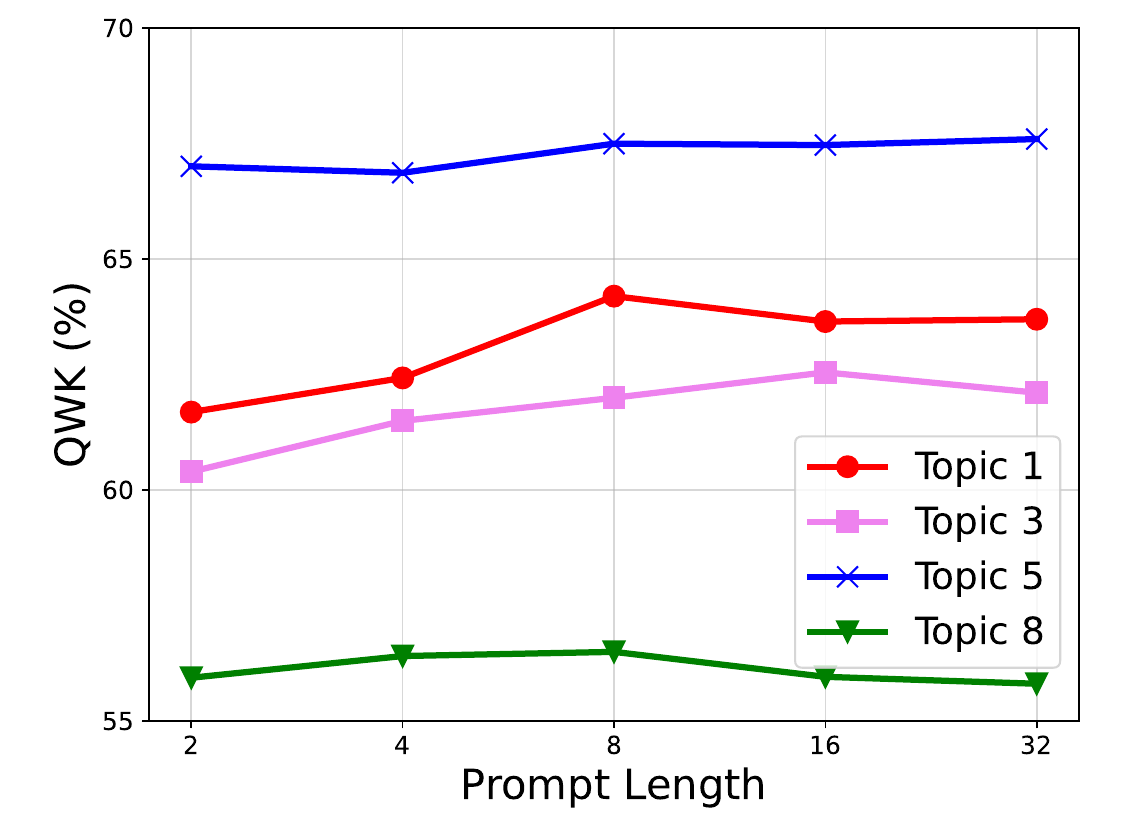}
}
% \subfloat[]{
% \includegraphics[width=0.3\linewidth]{figure/T8.pdf}
% }
\caption{The impact of hyperparameters on model performance.}
\vspace{-2mm}
\label{fig:hyper}
\end{figure}

 We investigated the impact of different hyperparameters on model performance, including loss weights and prompt length.  $\alpha$ and $\beta$ are trade-off parameters that balance the weights of the different losses. Given that regression and classification losses differ in magnitude, we employ a grid search strategy, sampling $\alpha$ from \{1, 5, 10, 50, 100\} and $\beta$ from \{0.1, 0.5, 1, 5, 10\}. As shown in Figure~\ref{fig:hyper} (a), the model's performance initially improves with increasing $\alpha$ and $\beta$, peaking at $\alpha=10$ and $\beta=1$, before declining. It is worth noting that there is a slight enhancement of QWK values at $\beta$ is 10 compared to 1, which may be caused by the dominance of adversarial loss. Additionally, despite the wide range of $\alpha$ and $\beta$ values, the model's performance remains relatively stable, demonstrating the robustness of the ATOP model across different parameter settings. Figure~\ref{fig:hyper} (b) illustrates the effect of prompt length on model performance with prompt lengths assumed to be identical for both topic-shared and topic-specific prompts, sampled from \{2, 4, 8, 16, 32\}. The results indicate that model performance is relatively insensitive to changes in prompt length, with optimal performance achieved using shorter prompts. Extending the prompt length does not result in significant performance gains.

% \begin{figure}[htp]\centering
%  \centering
%     \includegraphics[width=80mm]{figure/prompt length.pdf}
% \caption{Effect of the prompt length.}
% \vspace{-2mm}
% \label{fig:length}
% \end{figure}

\subsection{Parameter Efficiency}\label{sec:parameter}

Table 5 presents a comparison of parameter efficiency across different cross-topic frameworks, including Hi-Att, Bert-FT, PT-V2, and ATOP. As shown, both PT-V2 and ATOP—based on prompt tuning—significantly reduce the number of trainable parameters compared to full fine-tuning, while achieving superior performance. Notably, ATOP requires only 0.39\% of the trainable parameters used in full fine-tuning of BERT-base, matching the number of parameters of the classic Hi-Att framework, but substantially outperforming it in terms of average QWK.
 
 % Table ~\ref{table:parameter:parameter} compares the number of parameters of ATOP with baseline methods. As shown, ATOP significantly reduces the parameters compared to Bert-FT, while maintaining a comparable scale to the classical cross-topic scoring method Hi-att and the prompt-based method PT-V2. 

\begin{table}[t]\centering{
\caption{Comparison of parameter efficiency across different methods.} 
\begin{tabular}{lccccc}
\toprule
\multicolumn{2}{l}{Methods}        & Hi-att       & Bert-FT         & PT-V2      & ATOP  \\
\midrule

\multicolumn{2}{l}  {No. of parameters (M) } & 0.48  & {110}    & 0.61       & 0.43   \\
\multicolumn{2}{l}  {Average QWK} & 0.349 & {0.481}    & 0.542        & 0.594   \\
\bottomrule
\end{tabular}
}
\label{table:parameter:parameter}
\end{table}

\section{Conclusion}
% In this paper, we frame the learning of topic-shared and topic-specific features for the AES task as a joint optimization of topic-shared and topic-specific soft prompts within a prompt-tuning paradigm. To facilitate this, we incorporate adversarial training to learn the topic-shared prompt within a unified classification and regression framework. Additionally, we utilize a neighbor-based classifier to generate pseudo-labels for the target topic, which are subsequently used to refine the topic-specific prompt. Extensive experimental results demonstrate that our method outperforms existing SOTA methods, especially on some traits that are closely related to the topic. However, a limitation of our approach is that the topic-aware prompt is only incorporated at the input layer. In future work, we plan to extend the insertion of topic-aware prompts to each layer of the PLM to further improve the performance of cross-topic AES.
In this paper, we propose an Adversarial TOpic-aware Prompt-tuning (ATOP) method for cross-topic AES to address the limitations of traditional methods that primarily extract topic-shared features while overlooking topic-specific characteristics. ATOP formulates the learning of topic-shared and topic-specific features as a joint optimization of corresponding soft prompts within the prompt-tuning framework. Adversarial training between the topic-shared prompt and a topic discriminator is introduced to facilitate the learning of topic-specific prompts. Additionally, a pseudo-labeling strategy is employed to assign pseudo-labels to the target topic, further aiding in the learning of topic-specific prompts. Extensive experimental results demonstrate that ATOP outperforms existing state-of-the-art methods, particularly on traits that are strongly influenced by topic relevance.

\newpage
\begin{ack}
This work was supported by the National Natural Science Foundation of China (Grant Nos. 62077033 and 62302264), and by the Taishan Scholar Program of Shandong Province (Grant Nos. tsqn202211199 ).
\end{ack}
\normalem
\bibliography{m974}

\begin{thebibliography}{39}
\providecommand{\natexlab}[1]{#1}
\providecommand{\url}[1]{\texttt{#1}}
\expandafter\ifx\csname urlstyle\endcsname\relax
  \providecommand{\doi}[1]{doi: #1}\else
  \providecommand{\doi}{doi: \begingroup \urlstyle{rm}\Url}\fi

\bibitem[Attali and Burstein(2006)]{attali2006automated}
Y.~Attali and J.~Burstein.
\newblock Automated essay scoring with e-rater{\textregistered} v. 2.
\newblock \emph{The Journal of Technology, Learning and Assessment}, 4\penalty0 (3), 2006.

\bibitem[Brown et~al.(2020)Brown, Mann, Ryder, Subbiah, Kaplan, Dhariwal, Neelakantan, Shyam, Sastry, Askell, et~al.]{brown2020language}
T.~Brown, B.~Mann, N.~Ryder, M.~Subbiah, J.~D. Kaplan, P.~Dhariwal, A.~Neelakantan, P.~Shyam, G.~Sastry, A.~Askell, et~al.
\newblock Language models are few-shot learners.
\newblock \emph{Advances in neural information processing systems}, 33:\penalty0 1877--1901, 2020.

\bibitem[Cao et~al.(2020)Cao, Jin, Wan, and Yu]{cao2020domain}
Y.~Cao, H.~Jin, X.~Wan, and Z.~Yu.
\newblock Domain-adaptive neural automated essay scoring.
\newblock In \emph{Proceedings of the 43rd international ACM SIGIR conference on research and development in information retrieval}, pages 1011--1020, 2020.

\bibitem[Chen et~al.(2021)Chen, Wang, Wang, and Long]{chen2021representation}
X.~Chen, S.~Wang, J.~Wang, and M.~Long.
\newblock Representation subspace distance for domain adaptation regression.
\newblock In \emph{ICML}, pages 1749--1759, 2021.

\bibitem[Chen and Li(2023)]{chen2023pmaes}
Y.~Chen and X.~Li.
\newblock Pmaes: Prompt-mapping contrastive learning for cross-prompt automated essay scoring.
\newblock In \emph{Proceedings of the 61st Annual Meeting of the Association for Computational Linguistics (Volume 1: Long Papers)}, pages 1489--1503, 2023.

\bibitem[Chen and Li(2024)]{chen2024plaes}
Y.~Chen and X.~Li.
\newblock Plaes: Prompt-generalized and level-aware learning framework for cross-prompt automated essay scoring.
\newblock In \emph{Proceedings of the 2024 Joint International Conference on Computational Linguistics, Language Resources and Evaluation (LREC-COLING 2024)}, pages 12775--12786, 2024.

\bibitem[Devlin et~al.(2019)Devlin, Chang, Lee, and Toutanova]{devlin2019bert}
J.~Devlin, M.-W. Chang, K.~Lee, and K.~Toutanova.
\newblock Bert: Pre-training of deep bidirectional transformers for language understanding.
\newblock In \emph{Proceedings of the 2019 Conference of the North American Chapter of the Association for Computational Linguistics: Human Language Technologies, Volume 1 (Long and Short Papers)}, pages 4171--4186, 2019.

\bibitem[Do et~al.(2023)Do, Kim, and Lee]{do2023prompt}
H.~Do, Y.~Kim, and G.~G. Lee.
\newblock Prompt-and trait relation-aware cross-prompt essay trait scoring.
\newblock In \emph{Findings of the Association for Computational Linguistics: ACL 2023}, pages 1538--1551, 2023.

\bibitem[Donahue et~al.(2014)Donahue, Jia, Vinyals, Hoffman, Zhang, Tzeng, and Darrell]{donahue2014decaf}
J.~Donahue, Y.~Jia, O.~Vinyals, J.~Hoffman, N.~Zhang, E.~Tzeng, and T.~Darrell.
\newblock Decaf: A deep convolutional activation feature for generic visual recognition.
\newblock In \emph{International conference on machine learning}, pages 647--655. PMLR, 2014.

\bibitem[Dong et~al.(2017)Dong, Zhang, and Yang]{dong2017attention}
F.~Dong, Y.~Zhang, and J.~Yang.
\newblock Attention-based recurrent convolutional neural network for automatic essay scoring.
\newblock In \emph{Proceedings of the 21st conference on computational natural language learning (CoNLL 2017)}, pages 153--162, 2017.

\bibitem[Ganin and Lempitsky(2015)]{ganin2015unsupervised}
Y.~Ganin and V.~Lempitsky.
\newblock Unsupervised domain adaptation by backpropagation.
\newblock In \emph{International conference on machine learning}, pages 1180--1189. PMLR, 2015.

\bibitem[Hussein et~al.(2020)Hussein, Hassan, and Nassef]{hussein2020trait}
M.~A. Hussein, H.~A. Hassan, and M.~Nassef.
\newblock A trait-based deep learning automated essay scoring system with adaptive feedback.
\newblock \emph{International Journal of Advanced Computer Science and Applications}, 11\penalty0 (5), 2020.

\bibitem[Jiang et~al.(2023)Jiang, Gao, Yin, Liu, Yu, Cheng, and Gu]{jiang2023improving}
Z.~Jiang, T.~Gao, Y.~Yin, M.~Liu, H.~Yu, Z.~Cheng, and Q.~Gu.
\newblock Improving domain generalization for prompt-aware essay scoring via disentangled representation learning.
\newblock In \emph{Proceedings of the 61st Annual Meeting of the Association for Computational Linguistics (Volume 1: Long Papers)}, pages 12456--12470, 2023.

\bibitem[Jin et~al.(2018)Jin, He, Hui, and Sun]{jin2018tdnn}
C.~Jin, B.~He, K.~Hui, and L.~Sun.
\newblock Tdnn: a two-stage deep neural network for prompt-independent automated essay scoring.
\newblock In \emph{Proceedings of the 56th Annual Meeting of the Association for Computational Linguistics (Volume 1: Long Papers)}, pages 1088--1097, 2018.

\bibitem[Ke and Ng(2019)]{ke2019automated}
Z.~Ke and V.~Ng.
\newblock Automated essay scoring: A survey of the state of the art.
\newblock In \emph{Proceedings of the Twenty-Eighth International Joint Conference on Artificial Intelligence}, volume~19, pages 6300--6308, 2019.

\bibitem[Klebanov et~al.(2013)Klebanov, Madnani, and Burstein]{klebanov2013using}
B.~B. Klebanov, N.~Madnani, and J.~Burstein.
\newblock Using pivot-based paraphrasing and sentiment profiles to improve a subjectivity lexicon for essay data.
\newblock \emph{Transactions of the Association for Computational Linguistics}, 1:\penalty0 99--110, 2013.

\bibitem[Lester et~al.(2021)Lester, Al-Rfou, and Constant]{lester2021power}
B.~Lester, R.~Al-Rfou, and N.~Constant.
\newblock The power of scale for parameter-efficient prompt tuning.
\newblock In \emph{Proceedings of the 2021 Conference on Empirical Methods in Natural Language Processing}, pages 3045--3059, 2021.

\bibitem[Li and Chen(2024)]{li2024plaes}
X.~Li and Y.~Chen.
\newblock Plaes: Prompt-generalized and level-aware learning framework for cross-prompt automated essay scoring.
\newblock In \emph{Proceedings of the 2024 Joint International Conference on Computational Linguistics, Language Resources and Evaluation (LREC-COLING 2024)}, pages 12775--12786, 2024.

\bibitem[Li et~al.(2020)Li, Chen, and Nie]{li2020sednn}
X.~Li, M.~Chen, and J.-Y. Nie.
\newblock Sednn: Shared and enhanced deep neural network model for cross-prompt automated essay scoring.
\newblock \emph{Knowledge-Based Systems}, 210:\penalty0 106491, 2020.

\bibitem[Li and Liang(2021)]{li2021prefix}
X.~L. Li and P.~Liang.
\newblock Prefix-tuning: Optimizing continuous prompts for generation.
\newblock In \emph{Proceedings of the 59th Annual Meeting of the Association for Computational Linguistics and the 11th International Joint Conference on Natural Language Processing (Volume 1: Long Papers)}, pages 4582--4597, 2021.

\bibitem[Liang et~al.(2021)Liang, Hu, and Feng]{liang2021domain}
J.~Liang, D.~Hu, and J.~Feng.
\newblock Domain adaptation with auxiliary target domain-oriented classifier.
\newblock In \emph{Proceedings of the IEEE/CVF conference on computer vision and pattern recognition}, pages 16632--16642, 2021.

\bibitem[Liu et~al.(2021)Liu, Ji, Fu, Tam, Du, Yang, and Tang]{liu2021p}
X.~Liu, K.~Ji, Y.~Fu, W.~L. Tam, Z.~Du, Z.~Yang, and J.~Tang.
\newblock P-tuning v2: Prompt tuning can be comparable to fine-tuning universally across scales and tasks.
\newblock \emph{arXiv preprint arXiv:2110.07602}, 2021.

\bibitem[Mathias and Bhattacharyya(2018)]{mathias2018asap++}
S.~Mathias and P.~Bhattacharyya.
\newblock Asap++: Enriching the asap automated essay grading dataset with essay attribute scores.
\newblock In \emph{Proceedings of the eleventh international conference on language resources and evaluation (LREC 2018)}, 2018.

\bibitem[Mayfield and Black(2020)]{mayfield2020should}
E.~Mayfield and A.~W. Black.
\newblock Should you fine-tune bert for automated essay scoring?
\newblock In \emph{Proceedings of the Fifteenth Workshop on Innovative Use of NLP for Building Educational Applications}, pages 151--162, 2020.

\bibitem[Ramesh and Sanampudi(2022)]{ramesh2022automated}
D.~Ramesh and S.~K. Sanampudi.
\newblock An automated essay scoring systems: a systematic literature review.
\newblock \emph{Artificial Intelligence Review}, 55\penalty0 (3):\penalty0 2495--2527, 2022.

\bibitem[Ridley et~al.(2020)Ridley, He, Dai, Huang, and Chen]{ridley2020prompt}
R.~Ridley, L.~He, X.~Dai, S.~Huang, and J.~Chen.
\newblock Prompt agnostic essay scorer: a domain generalization approach to cross-prompt automated essay scoring.
\newblock \emph{arXiv preprint arXiv:2008.01441}, 2020.

\bibitem[Ridley et~al.(2021)Ridley, He, Dai, Huang, and Chen]{ridley2021automated}
R.~Ridley, L.~He, X.-y. Dai, S.~Huang, and J.~Chen.
\newblock Automated cross-prompt scoring of essay traits.
\newblock In \emph{Proceedings of the AAAI conference on artificial intelligence}, volume~35, pages 13745--13753, 2021.

\bibitem[Rodriguez et~al.(2019)Rodriguez, Jafari, and Ormerod]{rodriguez2019language}
P.~U. Rodriguez, A.~Jafari, and C.~M. Ormerod.
\newblock Language models and automated essay scoring.
\newblock \emph{arXiv preprint arXiv:1909.09482}, 2019.

\bibitem[Schick and Sch{\"u}tze(2021)]{schick2021exploiting}
T.~Schick and H.~Sch{\"u}tze.
\newblock Exploiting cloze-questions for few-shot text classification and natural language inference.
\newblock In \emph{Proceedings of the 16th Conference of the European Chapter of the Association for Computational Linguistics: Main Volume}, pages 255--269, 2021.

\bibitem[Sethi and Singh(2022)]{sethi2022natural}
A.~Sethi and K.~Singh.
\newblock Natural language processing based automated essay scoring with parameter-efficient transformer approach.
\newblock In \emph{2022 6th International Conference on Computing Methodologies and Communication (ICCMC)}, pages 749--756. IEEE, 2022.

\bibitem[Stahl et~al.(2024)Stahl, Biermann, Nehring, and Wachsmuth]{stahl2024exploring}
M.~Stahl, L.~Biermann, A.~Nehring, and H.~Wachsmuth.
\newblock Exploring llm prompting strategies for joint essay scoring and feedback generation.
\newblock In \emph{Proceedings of the 19th Workshop on Innovative Use of NLP for Building Educational Applications (BEA 2024)}, pages 283--298, 2024.

\bibitem[Taghipour and Ng(2016)]{taghipour2016neural}
K.~Taghipour and H.~T. Ng.
\newblock A neural approach to automated essay scoring.
\newblock In \emph{Proceedings of the 2016 conference on empirical methods in natural language processing}, pages 1882--1891, 2016.

\bibitem[Tay et~al.(2018)Tay, Phan, Tuan, and Hui]{tay2018skipflow}
Y.~Tay, M.~Phan, L.~A. Tuan, and S.~C. Hui.
\newblock Skipflow: Incorporating neural coherence features for end-to-end automatic text scoring.
\newblock In \emph{Proceedings of the AAAI conference on artificial intelligence}, volume~32, 2018.

\bibitem[Touvron et~al.(2023)Touvron, Lavril, Izacard, Martinet, Lachaux, Lacroix, Rozi{\`e}re, Goyal, Hambro, Azhar, et~al.]{touvron2023llama}
H.~Touvron, T.~Lavril, G.~Izacard, X.~Martinet, M.-A. Lachaux, T.~Lacroix, B.~Rozi{\`e}re, N.~Goyal, E.~Hambro, F.~Azhar, et~al.
\newblock Llama: Open and efficient foundation language models.
\newblock \emph{arXiv preprint arXiv:2302.13971}, 2023.

\bibitem[Vajjala(2018)]{vajjala2018automated}
S.~Vajjala.
\newblock Automated assessment of non-native learner essays: Investigating the role of linguistic features.
\newblock \emph{International Journal of Artificial Intelligence in Education}, 28:\penalty0 79--105, 2018.

\bibitem[Vaswani et~al.(2017)Vaswani, Shazeer, Parmar, Uszkoreit, Jones, Gomez, Kaiser, and Polosukhin]{vaswani2017attention}
A.~Vaswani, N.~Shazeer, N.~Parmar, J.~Uszkoreit, L.~Jones, A.~N. Gomez, {\L}.~Kaiser, and I.~Polosukhin.
\newblock Attention is all you need.
\newblock \emph{Advances in neural information processing systems}, 30, 2017.

\bibitem[Wu and Shi(2022)]{wu2022adversarial}
H.~Wu and X.~Shi.
\newblock Adversarial soft prompt tuning for cross-domain sentiment analysis.
\newblock In \emph{Proceedings of the 60th Annual Meeting of the Association for Computational Linguistics (Volume 1: Long Papers)}, pages 2438--2447, 2022.

\bibitem[Yang et~al.(2020)Yang, Cao, Wen, Wu, and He]{yang2020enhancing}
R.~Yang, J.~Cao, Z.~Wen, Y.~Wu, and X.~He.
\newblock Enhancing automated essay scoring performance via fine-tuning pre-trained language models with combination of regression and ranking.
\newblock In \emph{Findings of the Association for Computational Linguistics: EMNLP 2020}, pages 1560--1569, 2020.

\bibitem[Zhang et~al.(2025)Zhang, Deng, Dong, Zhao, Liu, and Cui]{zhang2025pairwise}
C.~Zhang, J.~Deng, X.~Dong, H.~Zhao, K.~Liu, and C.~Cui.
\newblock Pairwise dual-level alignment for cross-prompt automated essay scoring.
\newblock \emph{Expert Systems with Applications}, 265:\penalty0 125924, 2025.

\end{thebibliography}

\newpage
\appendix
\setcounter{table}{0} 
\renewcommand{\thetable}{\arabic{table}}
\section{Statistics of Datasets}

The ASAP++ dataset comprises 12,987 essays across 8 different topics and encompasses three genres: narrative, argumentative, and source-dependent. In the classification task, we divided essay scores into four categories. Specifically, the essay scores are first normalized to the range [0, 1]. Then we assign scores in the intervals (0, 0.4), (0.4, 0.6), (0.6, 0.8), and (0.8, 1) to their corresponding grades (excellent, good, moderate, and poor), respectively. We evaluated on the ASAP++ dataset, which builds on the original ASAP dataset with trait scores for all topics. Meanwhile, some traits only exists in separate topics, and the model could not learn relevant information from the source topic when these topics serve as target topics, so we removed these traits, such as the style trait of prompt 7 and the sound trait of prompt 8. Table 1 displays the statistics for the ASAP++ datasets.

\begin{table*}[htbp]\centering{
\caption{Detailed statistics of the ASAP++ datasets. Cont: Content, Org: Organization, WC: Word Choice, SF: Sentence Fluency, Conv: Conventions, TA: Topic Adherence, Lan: Language, Nar: Narrativity.}
\begin{tabular*}{\hsize}{@{}@{\extracolsep{\fill}}cccccccccc@{}}
\toprule
Topic & Essays & Genre & Traits & Length & Score & Poor & Moderate & Good & Excellent \\
\midrule 
1& 1783 & Arg &Cont, Org, WC, SF, Conv & 350 & 2-12 & 45 & 245 & 1021 & 472\\
2& 1800 & Arg &Cont, Org, WC, SF, Conv & 350 & 1-6  & 177 & 763 & 778 & 82\\
3& 1726 &  Sou &Cont, TA, Lan, Nar & 150 & 0-3  & 39 & 607 & 657 & 423 \\
4& 1772 &  Sou &Cont, TA, Lan, Nar & 150 & 0-3  & 312 & 636 & 570 & 254 \\
5& 1805 &  Sou &Cont, TA, Lan, Nar & 150 & 0-4  & 326 & 649 & 572 & 258 \\
6& 1800 &  Sou &Cont, TA, Lan, Nar & 150 & 0-4  & 211 & 405 & 817 & 367 \\
7& 1569 &  Nar &Cont, Org, Conv & 250 & 0-30 & 268 & 717 & 488 & 96 \\
8& 723 &  Nar &Cont, Org, WC, SF, Conv & 650 & 0-60  & 9 & 275 & 418 & 21 \\
\bottomrule
% \end{tabular}
\end{tabular*}
}
% }
\label{table:ASAP++}
\end{table*}

\begin{table*}[htbp]\centering
\caption{QWK scores of the holistic score classification task for each topic.}
 \begin{tabular*}{\textwidth}{@{}@{\extracolsep{\fill}}lccccccccccc@{}}
\toprule
\multicolumn{2}{l}{Methods}              & Topic 1         & Topic 2         & Topic 3         & Topic 4         & Topic 5         & Topic 6         & Topic 7         & Topic 8         & Avg              \\
\midrule

\multicolumn{2}{l}{Hi-att}                 & 0.311          & 0.423          & 0.459         & 0.568         & 0.517          & 0.560         & 0.414        & 0.241         & 0.437         \\
\multicolumn{2}{l}{Bert-FT}                  & 0.493         & 0.568         & 0.475          & 0.613          & 0.649          & 0.578         &  0.496         & 0.336          & 0.526          \\
\multicolumn{2}{l}{PT-V2}       & 0.517         & 0.608        & 0.673        & 0.754          & 0.670         & 0.598         & 0.641         & 0.463            & 0.617        \\
\multicolumn{2}{l}{PAES}                & 0.529         & 0.595        & 0.547        & 0.668          & 0.659         & 0.552         & 0.542         & 0.357         & 0.590          \\
\multicolumn{2}{l}{PMAES}                & \uline{0.534}          & \uline{0.624}         &\uline{0.631}         & \uline{0.710}         & \uline{0.697}       & \uline{0.603}         & \uline{0.553}         & \uline{0.385}         & \uline{0.592}       \\
\multicolumn{2}{l}{ATOP (Ours)}                 & \textbf{0.557}    & \textbf{0.632} & \textbf{0.663} & \textbf{0.752}    & \textbf{0.729} & \textbf{0.634}    & \textbf{0.590}    & \textbf{0.449}    & \textbf{0.626}   \\
\bottomrule
% \end{tabular}
\end{tabular*}
% }
\label{table:classfication}
\end{table*}

% \begin{table*}[hb]
% \setlength{\tabcolsep}{1mm}{
% \renewcommand{\arraystretch}{1}
% \begin{tabular}{ccccccccccc}
% \toprule
% \textbf{Topic} & \textbf{Essays} & \textbf{Genre} & \textbf{Traits} & \textbf{Mean Length} & \textbf{Score Range} & \textbf{Poor} & \textbf{Moderate} & \textbf{Good} & \textbf{Excellent} \\
% \midrule 
% 1& 1783 & Arg &Cont, Org, WC, SF, Conv & 350 & 2-12 & 45 & 245 & 1021 & 472\\
% 2& 1800 & Arg &Cont, Org, WC, SF, Conv & 350 & 1-6  & 177 & 763 & 778 & 82\\
% 3& 1726 &  Sou &Cont, TA, Lan, Nar & 150 & 0-3  & 39 & 607 & 657 & 423 \\
% 4& 1772 &  Sou &Cont, TA, Lan, Nar & 150 & 0-3  & 312 & 636 & 570 & 254 \\
% 5& 1805 &  Sou &Cont, TA, Lan, Nar & 150 & 0-4  & 326 & 649 & 572 & 258 \\
% 6& 1800 &  Sou &Cont, TA, Lan, Nar & 150 & 0-4  & 211 & 405 & 817 & 367 \\
% 7& 1569 &  Nar &Cont, Org, Conv & 250 & 0-30 & 268 & 717 & 488 & 96 \\
% 8& 723 &  Nar &Cont, Org, WC, SF, Conv & 650 & 0-60  & 9 & 275 & 418 & 21 \\
% \bottomrule
% \end{tabular}
% }
% \vspace{-2mm}
% \caption{Detailed statistics of the ASAP++ datasets. Cont: Content, Org: Organization, WC: Word Choice, SF: Sentence Fluency, Conv: Conventions, TA: Topic Adherence, Lan: Language, Nar: Narrativity.}
% \label{table:ASAP++}
% \end{table*}
\begin{table*}[htbp]\centering
\caption{QWK scores of the holistic score regression task for each topic. $\star$ refers to the results from~\cite{stahl2024exploring}.}
\begin{tabular*}{\hsize}{@{}@{\extracolsep{\fill}}lccccccccccc@{}}
\toprule
\multicolumn{2}{l}{Methods}              & Topic 1         & Topic 2         & Topic 3         & Topic 4         & Topic 5         & Topic 6         & Topic 7         & Topic 8         & Avg              \\
\midrule

\multicolumn{2}{l}{Hi-att}                 & 0.372          & 0.465          & 0.432         & 0.523         & 0.586          & 0.574         & 0.514        & 0.323         & 0.474         \\
\multicolumn{2}{l}{Bert-FT}                  & 0.561         & 0.528         & 0.493          & 0.574          & 0.688          & 0.587         &  0.605         & 0.440          & 0.560          \\
\multicolumn{2}{l}{PT-V2}       & 0.651         & 0.614        & 0.656        & 0.638          & 0.685         & 0.589         & 0.641         & 0.463            & 0.617        \\
\multicolumn{2}{l}{PAES}                & {0.677}         &  0.596        & 0.564        & 0.542          & 0.673         & 0.562         & 0.629         & 0.421         & 0.583          \\
\multicolumn{2}{l}{PMAES}                &  \uline{0.684}          &  \uline{0.628}         &\uline{0.609}         & {0.575}         & \uline{0.713}       & {0.591}         & \uline{0.668}         & \uline{0.503}         & \uline{0.621}       \\
\multicolumn{2}{l}{${LLM-fs}^{\star}$}&0.448&0.585&0.479&\uline{0.596}&0.557&\textbf{0.649}&0.438&0.481&0.529    \\
\multicolumn{2}{l}{ATOP (Ours)}                 & \textbf{0.692}    & \textbf{0.645} & \textbf{0.648} & \textbf{0.632}    & \textbf{0.754} & \uline{0.611}    & \textbf{0.701}    & \textbf{0.544}    & \textbf{0.653}   \\
\bottomrule
% \end{tabular}
\end{tabular*}
% }
\label{table:regression}
\end{table*}
\begin{figure*}[htbp]
\centering
\subfloat[]{
\includegraphics[width=0.3\linewidth]{T1.pdf}
}
\subfloat[]{
\includegraphics[width=0.3\linewidth]{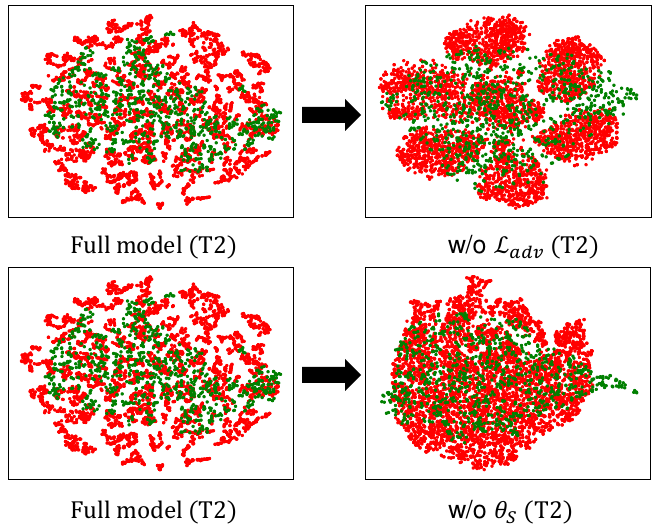}
}
\subfloat[]{
\includegraphics[width=0.3\linewidth]{T3.pdf}
}\\
\subfloat[]{
\includegraphics[width=0.3\linewidth]{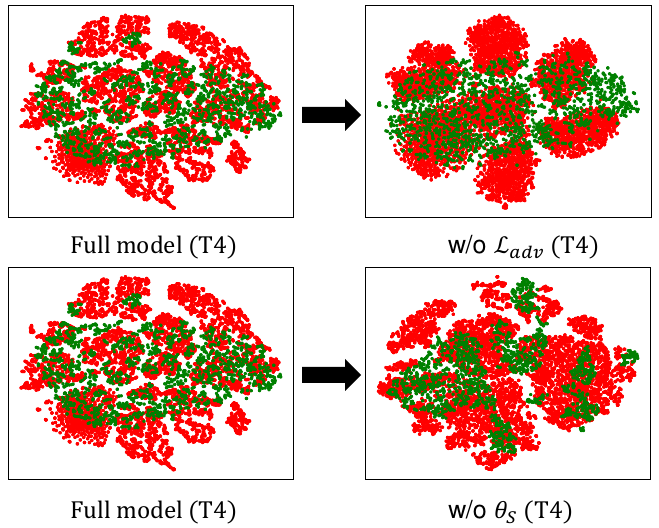}
}
\subfloat[]{
\includegraphics[width=0.3\linewidth]{T5.pdf}
}
\subfloat[]{
\includegraphics[width=0.3\linewidth]{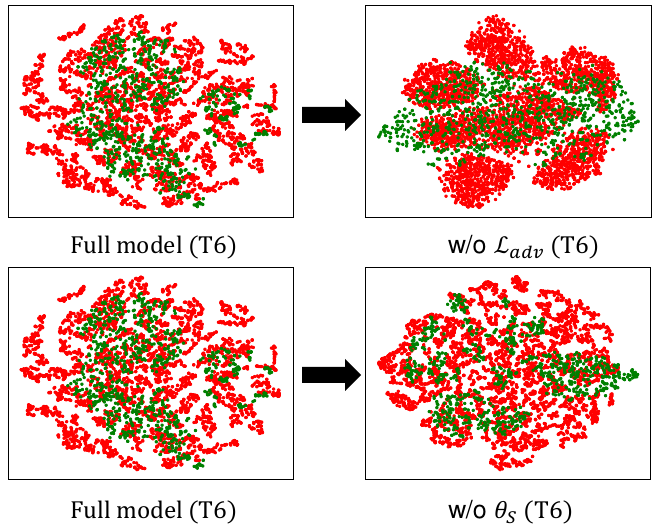}
}\\
\subfloat[]{
\includegraphics[width=0.3\linewidth]{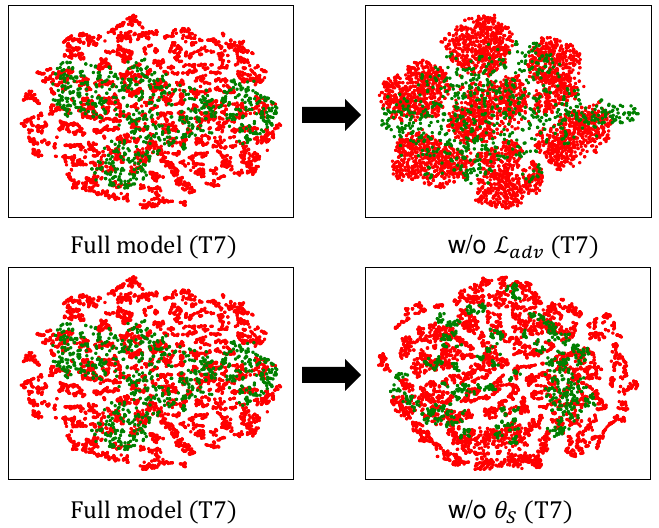}
}
\subfloat[]{
\includegraphics[width=0.3\linewidth]{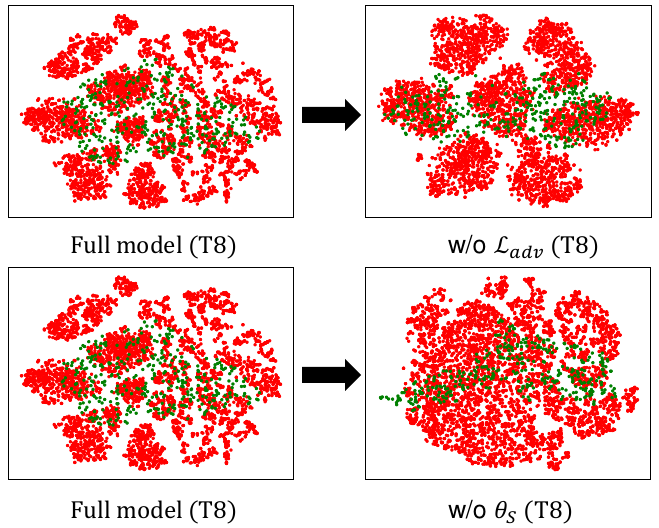}
}
\caption{These figures show the visualization results for each topic, respectively. Red represents the features of the source topic, green represents the features of the target topic.}
\label{fig:visual}
\end{figure*}

\section{Implementation Details}

For all of our experiments, we use Bert-base containing a 12-layer Transformer as our pre-trained language model(PLM), and tokenize all the words via its tokenizer. The output feature dimension of PLM is 768 dimensions, which is reduced to 100 dimensions by a linear layer, and then concatenated with 86 dimensions of handcrafted features containing information on length, text complexity, sentiment and text variation. We set one head and input embedding dimension of 186 for the trait attention. The input dimension of the MLP is 186, the intermediate dimension is 10, and the output dimension is 4. For regression tasks, we perform inverse normalization on the final regression results of the model to restore them to the original score range.

During training, the model is trained for 30 epochs with batch size of 4 for each topic. The loss trade-off parameters $\alpha$ and $\beta$ are 10 and 1, respectively. The prompt lengths $m$ and $n$ are both 8.
We optimized the model parameters using the Adam optimizer with an initialized learning rate of 0.01, a decay rate of 0.9, and a decay step of 2000. For the multi-trait AES task, we treat all traits as equally important and take the epoch with the highest average QWK for all traits as the best result. The code is implemented in Pytorch 1.9.0 and Python 3.7.5 and train on 4 Nvidia Gefore GTX TITAN GPUs.

\section{Baselines}
To evaluate the effectiveness of our proposed model, we compared it against eight state-of-the-art methods across three tasks: holistic score classification, holistic score regression, and multi-trait score regression. Baseline models include Hi-att~\cite{dong2017attention}, AES aug~\cite{hussein2020trait}, Bert-FT~\cite{devlin2019bert}, PAES~\cite{ridley2020prompt}, CTS~\cite{ridley2021automated}, PMAES~\cite{chen2023pmaes}, PLAES~\cite{li2024plaes}, and LLM-fs~\cite{stahl2024exploring}. Note that since LLM-fs performs only holistic score regression for essays, we report its experimental results solely for the overall regression task.
\begin{itemize}
    \item Hi-ATT \cite{dong2017attention} introduces a CNN-LSTM-ATT hierarchical architecture designed to capture feature representations at the word, sentence, and essay levels. As a representative model in AES, Hi-ATT remains a foundational framework, serving as the basis for several advanced AES models in recent research \cite{ridley2020prompt,ridley2021automated,chen2023pmaes}.
    % \item Hi-att \cite {dong2017attention} utilizes the CNN-LSTM-ATT architecture to model feature representations of essays. 

    \item AES aug \cite{hussein2020trait} is a multi-trait AES model built upon the framework proposed in \cite{taghipour2016neural}. It extends the original architecture by incorporating separate output linear layers for each trait. 
    % \item AES aug \cite{hussein2020trait} extends the model proposed in \cite{taghipour2016neural} to a multi-task framework.
    \item Bert-FT \cite{devlin2019bert} is a fine-tuned BERT-based model using all original topic essays to perform essay scoring and rating classification.

    % \item PAES \cite{ridley2020prompt} introduces handcrafted features into Hi-att model for holistic scoring .
    \item PAES \cite{ridley2020prompt} is a powerful holistic cross-topic scoring model that extends the Hi-att model architecture and introduces hand-crafted features into a cross-topic AES task.

    % \item CTS \cite{ridley2021automated} proposed a model specialized for cross-topic multi-trait essay scoring based on PAES, which designed shared layers to learn the word-level representation and sentence-level representation of the essay. For each trait, a separate private layer learning essay representation is designed. Meanwhile, the model considers the relevance of different traits and realizes the information interaction between the private layers of each trait via the attention mechanism.
    \item CTS \cite{ridley2021automated}  extends the PAES framework into a unified scoring model encompassing both holistic and trait-based assessment by incorporating an attention mechanism designed to model the interrelationships among various traits.

    % \item PMAES \cite{chen2023pmaes} introduces contrastive learning to the cross-topic trait AES task to learn more consistent features from different topics, which designs a topic-mapping contrastive learning strategy that treats the representation of the essay in the source topic and its weighted projected representation in the target topic as positive pairs, and treats other samples within the same batch as negative pairs for contrastive learning.  
    \item PMAES \cite{chen2023pmaes} introduces topic mapping to learn consistent representations of source and target topics for holistic and multi-trait essay scoring. 
     
    \item PLAES \cite{li2024plaes} trains a generalized AES model via meta-learning to capture generalized knowledge across topics, and designs a level-aware module based on contrastive learning to enhance the model's ability to differentiate writing quality.
    \item{{LLM-fs}}~\cite{stahl2024exploring} explores the use of different text prompt templates to guide the large language model Mistral in scoring essays and generating feedback. 
    
\end{itemize}

\section{Experimental Results}

% To demonstrate the superiority of ATOP on different tasks, we compared it with existing models on holistic score classification task and holistic score regression task.

% For the holistic score classification task and holistic score regression task, Table~\ref{table:classfication} shows that our method achieves the best result and outperforms the optimal baseline model by 2.7$\%$ in terms of average QWK on all topics, which demonstrates the effectiveness and advancement of our method on this task. Our experimental results from Table~\ref{table:classfication} show the poor performance of the Hi-att model, which indicates that there are significant domain differences between the source and target topics, and that directly transferring the model trained on the source topic data to the target topic leads to a significant performance degradation. In addition, three improved methods (PAES, CTS and PMAES) based on the Hi-att model have achieved better classification results, which shows that adding generic features of essays (e.g., POS features and handcrafted features) can improve the generalization of the model.

% Table~\ref{table:regression} lists the performance of our method on the holistic score regression task with the corresponding baseline methods. As with the results on the classification task, the average QWK score of our method on the regression task exceeds other baseline methods, which further confirms the effectiveness of our method on different tasks.

To further illustrate ATOP’s advantages across various tasks, we present comparative experimental results for holistic score classification and regression in Table~\ref{table:classfication} and Table~\ref{table:regression}. As demonstrated, our model outperforms the best baseline models in terms of average QWK values across all topics, achieving improvements of 3.4\% and 3.2\%, respectively. These results highlight the robustness and effectiveness of our model in both holistic score classification and regression tasks. Notably, as shown in Table~\ref{table:regression}, the performance of LLM-fs is significantly lower than that of existing cross-topic AES approaches. This underperformance can be primarily attributed to the inherent complexity of AES, coupled with the limitations of LLM-fs's complete reliance on manually designed text prompt templates to elicit LLMs' scoring ability. This disparity suggests that effectively applying LLMs to complex tasks such as AES necessitates the development of more sophisticated techniques and the integration of task-specific information to achieve competitive performance.

\section{Visualization}

To illustrate more visually the role of domain adversarial and topic-specific soft prompts, we visualize the feature representation of ATOP, ATOP w/o ${\mathcal{L}_{adv}}$ and ATOP w/o $\bm{\theta_{Sp}}$, respectively. As shown in Figure~\ref{fig:visual}, we supplemented the visualization in the main text and take each topic as the target topic respectively. 

As can be seen in Figure~\ref{fig:visual}, similar to topic 1 and topic 5, the other topics show the same trend. After removing the adversarial training, the feature representations of different source topic samples form distinct clusters. This indicates that adversarial training encourages the extraction of topic-agnostic features, as it reduces the model's ability to distinguish features by topic. Additionally, without adversarial training module, the feature distributions of the source and target topics become more separated, reflecting a reduction in cross-topic consistency.

Similarly, removing the topic-specific prompts leads to unconstrained learning of the target topic's feature distribution, resulting in scattered and disorganized representations. This highlights the crucial role of topic-specific prompts in guiding the model to learn well-structured and compact topic-specific feature representations, thereby preserving the coherence of the feature space.

\end{document}